\documentclass{article}

    \PassOptionsToPackage{numbers, compress}{natbib}

\usepackage[preprint]{neurips_2026}

\usepackage[utf8]{inputenc} 
\usepackage[T1]{fontenc}    
\usepackage{hyperref}       
\usepackage{url}            
\usepackage{booktabs}       
\usepackage{amsfonts}       
\usepackage{nicefrac}       
\usepackage{microtype}      
\usepackage{xcolor}         

\usepackage{amsmath}
\usepackage{amssymb}
\usepackage{mathtools}
\usepackage{amsthm}

\usepackage{multirow}
\usepackage{textcomp}
\usepackage{enumitem}
\usepackage{colortbl}
\usepackage{tabularx}
\usepackage{marvosym}
\usepackage[figuresright]{rotating}
\definecolor{LightGray}{gray}{0.96}
\definecolor{mydarkgreen}{RGB}{0, 150, 0}

\theoremstyle{plain}
\newtheorem{theorem}{Theorem}[section]

\newtheorem{lemma}[theorem]{Lemma}

\theoremstyle{definition}

\theoremstyle{remark}

\usepackage{wrapfig}

\title{FoMEMO: Towards Foundation Models for Expensive Multi-objective Optimization}

\author{%
  \textbf{Yiming Yao\textsuperscript{1,3}} \quad
  \textbf{Fei Liu\textsuperscript{1}} \quad
  \textbf{Liang Zhao\textsuperscript{1,3}} \quad
  \textbf{Xi Lin\textsuperscript{2}} \quad
  \textbf{Yilu Liu\textsuperscript{1,3}} \quad
  \textbf{Qingfu Zhang\textsuperscript{1,3}} \\
  \normalfont\textsuperscript{1}Department of Computer Science, City University of Hong Kong, Hong Kong, China \\
  \normalfont\textsuperscript{2}School of Mathematics and Statistics, Xi’an Jiaotong University, Xi’an, China \\
  \normalfont\textsuperscript{3}The City University of Hong Kong Shenzhen Research Institute, Shenzhen, China \\
  \normalfont\texttt{\{yimingyao3-c,fliu36-c,yilu.liu\}@my.cityu.edu.hk} \\
  \normalfont\texttt{xi.lin@xjtu.edu.cn, \{liang.zhao,qingfu.zhang\}@cityu.edu.hk} \\
}

\begin{document}

\maketitle

\begin{abstract}
Expensive multi-objective optimization is a prevalent and crucial concern in many real-world scenarios, where sample-efficiency is vital due to the limited evaluations to recover the true Pareto front for decision making. Existing works either involve repeatedly fitting Gaussian process surrogates from scratch for each newly encountered problem, or rely on computationally demanding hypervolume-oriented policy learning for amortized optimization, making it challenging to achieve scalable pre-training and efficient yet robust adaptation for diverse emerging real-world applications. To address these challenges, we propose \textbf{FoMEMO} (\textbf{Fo}undation \textbf{M}odels for \textbf{E}xpensive \textbf{M}ulti-objective \textbf{O}ptimization), which adopts a decomposition-based training paradigm that converts multi-objective optimization into preference-wise aggregated posterior learning tasks, enabling scalable pre-training on hundreds of millions of diverse synthetic datasets without relying on extensive real-world domain experiments. At test time, given observed trajectories from unseen problems and user preferences, the foundation model performs in-context posterior prediction and optimizes the derived acquisition functions for efficient candidate generation, achieving strong generalization and optimization performance without any subsequent model training or updates. Our code is available at {\hypersetup{hidelinks}\urlstyle{same}\textcolor{magenta}{\url{https://github.com/ymyao/FoMEMO}}}.


\end{abstract}

\section{Introduction}
\label{sec:introduction}
Optimizing multiple competitive objectives simultaneously is ubiquitous in many real-world applications, such as neural architecture search \cite{ying2019bench}, engineering design \cite{tanabe2020easy} and scientific discovery \cite{dara2022machine}.
The primary difficulty in these scenarios stems from the black-box nature of real-world evaluations, which are often costly and time-consuming. When decision-makers have access to only a limited number of evaluations, it becomes challenging to accurately recover the unknown true Pareto fronts.

To achieve sample-efficiency, traditional multi-objective Bayesian optimization (MOBO) typically employs Gaussian Process (GP) surrogates \cite{williams1995gaussian} to approximate objective functions, and then optimize acquisition functions to generate candidate solutions to update the surrogates iteratively in an online manner \cite{daulton2020differentiable,daulton2021parallel}. However, these methods typically need to rebuild surrogates from scratch when encountering a new problem, and the efficiency is greatly limited by the cost of GP training and inference \cite{williams2006gaussian}. To enhance the generalization and efficiency in downstream optimization, some research focuses on pre-training deep learning models in offline scenarios \cite{xue2024offline,yuan2024paretoflow}. These approaches typically assume that access to a large amount of real-world experimental data is available for training. However, acquiring large datasets in practical applications, especially in unknown or emerging fields, is often challenging or even impossible. While recent work has explored pre-training with synthetic functions for amortized multi-objective optimization in the online setting \cite{hung2025boformer,zhang2025context}, these methods mainly rely on hypervolume-oriented acquisition or query-policy learning. Such training objectives can become computationally demanding when scaled to massive synthetic data generation across diverse design or objective dimensions, trajectory lengths, and function properties, thereby limiting the diversity of the pre-training distribution and the robustness of generalization to unseen scenarios.

In real-world contexts, a wide range of multi-objective optimization problems arise daily across various fields, each exhibiting unique characteristics and necessitating rapid decision-making. Given the limitations of existing methodologies, a compelling research question arises: 

\textit{Can we develop a scalable foundation-model-style framework for expensive multi-objective optimization that adapts efficiently and robustly to a wide range of emerging real-world applications?}

In this paper, we attempt to answer this question by exploring a new paradigm named FoMEMO. It enables us to establish a foundation model conditioned on arbitrary domain trajectories and user preferences, and facilitates high-quality and efficient in-context optimization based on the predicted preference-wise aggregated posteriors. What distinguishes FoMEMO from existing methods lies in two aspects. First, unlike traditional MOBO methods that require repeatedly fitting surrogate models for each newly encountered problem and updating them during optimization, the foundation model in FoMEMO is pre-trained once and can be adapted to unseen scenarios in context without any subsequent model training or parameter updates throughout the entire optimization process. Second, unlike recent amortized RL-based methods that learn acquisition or query-generation policies, FoMEMO amortizes posterior reasoning while retaining a posterior-based optimization interface. It reformulates expensive multi-objective optimization as preference-wise aggregated posterior learning in a scalable supervised manner. By using decomposition-based scalarization, it can efficiently generate massive preference-indexed training signals from synthetic functions, avoiding the heavy cost of hypervolume-oriented training objectives when scaled to massive pre-training regimes. This makes our framework better suited for large-scale training data generation and thus better aligned with the foundation-model paradigm. The main contributions of this paper are as follows:

\begin{itemize}
\item We introduce a decomposition-based scalable pre-training paradigm for expensive multi-objective optimization. By converting multi-objective optimization into preference-wise aggregated posterior learning tasks, this paradigm enables efficient generation of massive supervised training signals from diverse synthetic functions, making our framework well suited for large-scale training-data generation and naturally aligned with the foundation-model paradigm.

\item We pre-train a foundation model once on hundreds of millions of synthetic datasets without relying on extensive real-world experimental data, enabling it to learn to predict preference-wise aggregated posteriors conditioned on observed optimization trajectories and user preferences. Based on the predicted posteriors, we derive both preference-based and preference-free acquisition functions, which can be optimized in context for efficient candidate generation without any subsequent model training or parameter updates.

\item We evaluate our method on a diverse set of synthetic benchmarks, engineering design problems, and real-world hyperparameter optimization tasks. Experimental results show that FoMEMO achieves strong generalization and robust optimization performance across unseen problems, outperforming existing MOBO and recent amortized baselines in the majority of cases while achieving a favorable candidate-generation efficiency.
\end{itemize}

\section{Preliminaries and Related Work}
\label{sec:preliminaries_related_work}

\subsection{Preliminaries}
\label{sec:preliminaries}
\paragraph{Expensive Multi-objective Optimization} In this paper, we address the following multi-objective optimization problem (MOP):
\begin{equation}
\min _{\boldsymbol{x} \in \mathcal{X}} \boldsymbol{f}(\boldsymbol{x})=(f_1(\boldsymbol{x}), f_2(\boldsymbol{x}), \cdots, f_m(\boldsymbol{x})),
\end{equation}
where $\boldsymbol{x}=(x_{1}, x_{2}, \ldots, x_{d})^T$ is the feature vector in the decision space $\mathcal{X} \subset \mathbb{R}^d$, and $\boldsymbol{f}: \mathcal{X} \rightarrow \mathbb{R}^m$ is the objective vector containing $m$ continuous objective functions. We consider the scenario in which all the objective functions are expensive to evaluate, with no known analytical expressions or gradient information. Typically, there is no single solution that can simultaneously achieve the optimal for all objectives. Instead, the goal is to find a set of solutions with Pareto optimal objective trade-offs. We say an objective vector $\boldsymbol{f}(\boldsymbol{x})$ 
Pareto dominates another $\boldsymbol{f}(\boldsymbol{x}^{\prime})$, denoted as $\boldsymbol{f}(\boldsymbol{x})\prec\boldsymbol{f}(\boldsymbol{x}^{\prime})$ if $f_{i}(\boldsymbol{x}) \leq f_{i}(\boldsymbol{x}^{\prime}), \forall i\in\{1, \ldots, m\}$ and there exists $j \in\{1, \ldots, m\}$ such that $ f_{j}(\boldsymbol{x}) < f_{j}(\boldsymbol{x}^{\prime})$. A solution $\boldsymbol{x}^*\in\mathcal{X}$ is Pareto optimal if there is no $\boldsymbol{x}$ such that $\boldsymbol{f}(\boldsymbol{x})\prec \boldsymbol{f}(\boldsymbol{x}^*)$. The set of all Pareto optimal solutions is called the Pareto set (PS), and the image of PS in the objective space is called the Pareto front (PF).

\paragraph{Scalarization for Multiple Objectives} Scalarization is a straightforward and effective technique for transforming multiple objectives into a single metric. A classic approach is based on decomposition, in which multiple objectives are aggregated using a family of scalarization functions $s_{\boldsymbol{\lambda}}(\boldsymbol{x}): \mathbb{R}^d \rightarrow \mathbb{R}$ parameterized by a set of preferences $\Delta=\{\boldsymbol{\lambda}\in\mathbb{R}_{+}^{m}\mid\|\boldsymbol{\lambda}\|_{1}=1\}$. We consider the popular Tchebycheff scalarization function:
\begin{equation}
s_{\boldsymbol{\lambda}}^{\text{tch}}(\boldsymbol{x})=\max _{1 \leq i \leq m}\{\lambda_i(y_i(\boldsymbol{x})-z_i^*)\},
\end{equation}
where $y_i(\boldsymbol{x})$ is the $i$-th observation value of the corresponding true function $f_i(\boldsymbol{x})$ at the input $\boldsymbol{x}$, $\boldsymbol{z}^*=(z_1^*, \cdots, z_m^*)$ is the ideal point (lower bound) in the objective space. A promising property behind Tchebycheff scalarization function is that we can identify each Pareto optimal solution by optimizing the scalarization function with a specific but unknown preference \cite{choo1983proper}:
\begin{equation}
\boldsymbol{x}^*=\arg \min _{\boldsymbol{x} \in \mathcal{X}} s_{\boldsymbol{\lambda}}^{\text{tch}}(\boldsymbol{x}).
\end{equation}
In our work, we use the negative of the scalarization function as the aggregation target $g_{\boldsymbol{\lambda}}(\boldsymbol{x})=-s_{\boldsymbol{\lambda}}(\boldsymbol{x})$ for the convenience of model training and inference. Thus, we wish to maximize $g_{\boldsymbol{\lambda}}(\boldsymbol{x})$ given each preference vector $\boldsymbol{\lambda}$.

Another representative scalarization method is the hypervolume (HV) indicator \cite{zitzler2002multiobjective}, which is monotonic to the Pareto dominance relation. The hypervolume of a finite approximate Pareto front $\mathcal{P}$ is defined as:
\begin{equation}
\text{HV}(\mathcal{P})=\boldsymbol{\Lambda}_m(\{\boldsymbol{v}\in\mathbb{R}^m|\exists\boldsymbol{p}\in \mathcal{P}:\boldsymbol{p}\prec\boldsymbol{v}\prec\boldsymbol{r}\}),
\label{eq:HV}
\end{equation}
where $\boldsymbol{\Lambda}_m$ denotes the m-dimensional Lebesgue measure, and $\boldsymbol{r}\in\mathbb{R}^m$ is a pre-defined reference point.

\paragraph{Bayesian Optimization} Bayesian optimization (BO) \cite{shahriari2015taking,frazier2018tutorial,garnett2023bayesian} is a sample-efficient technique for solving expensive black-box optimization problems. BO utilizes probabilistic surrogate models, typically Gaussian processes (GPs) \cite{williams2006gaussian}, to provide a prediction with calibrated uncertainty for the latent functions $\boldsymbol{f}$ based on the true observed data $D_n=\{\boldsymbol{x}_{i},\boldsymbol{y}_{i}\}_{i=1}^{n}$. To suggest the candidate solutions for next evaluations, BO defines and optimizes an acquisition function $\alpha: \mathcal{X} \mapsto \mathbb{R}$, which quantifies the utility value of evaluating a new set of solutions based on the predictive distribution of the surrogate models, effectively balancing exploration and exploitation.

\subsection{Related Work}
\label{sec:related_work}
\paragraph{Multi-objective Bayesian Optimization} Multi-objective Bayesian optimization (MOBO) typically employs GP models and acquisition functions to identify Pareto optimal solutions within a constrained evaluation budget. A popular category of MOBO is based on decomposition, wherein the original MOP is converted into several single-objective sub-problems that are easier to solve (e.g., ParEGO \cite{knowles2006parego}, TS-TCH \cite{paria2020flexible} and MOEA/D-EGO \cite{zhang2009expensive}). Besides decomposition, indicator-based approaches provide an integrated performance measure across all objectives under uncertainty. Prominent examples include EHVI \cite{emmerich2006single}, which generalizes the Expected Improvement (EI) function \cite{jones1998efficient} to multi-objective settings, along with its modern variants: the parallelizable qEHVI \cite{wada2019bayesian,daulton2020differentiable} and the noise-robust qNEHVI \cite{daulton2021parallel}. Another representative class stems from information theory, focusing on selecting points that maximize information gain regarding the Pareto set or front (e.g., PESMO \cite{hernandez2016predictive}, MESMO \cite{belakaria2019max} and JES \cite{tu2022joint,hvarfner2022joint}). In addition to model management and acquisition functions design, various batch sample selection strategies have also been developed in parallel setting (e.g., TSEMO \cite{bradford2018efficient}, USEMO-EI \cite{belakaria2020uncertainty} and DGEMO \cite{konakovic2020diversity}). While effective, these traditional MOBO methods typically require training individual models for every new problem from scratch. Furthermore, the substantial computational overhead associated with GP training and inference significantly hinders their scalability.

\paragraph{Learning-based Expensive Optimization} To enhance the generalization and scalability in downstream optimization, several approaches in the offline context involves training a deep neural network (DNN) on a static dataset to serve as an oracle \cite{trabucco2021conservative,yu2021roma,chen2023parallel}. This is followed by applying search methods, such as gradient descent or evolutionary algorithms, to generate a set of candidate solutions that are potentially superior to those present in the dataset \cite{xue2024offline,yuan2024paretoflow}. However, these approaches typically require a large amount of training data from expensive experiments, which is often impractical for various emerging applications in the real-world. Moreover, they often necessitate the training of dedicated models for specific problems, which limits their ability to generalize to unseen scenarios. To circumvent the need for extensive real-world training data, inspired by meta-learning frameworks in single-objective BO \cite{volpp2019meta,hsieh2021reinforced,maraval2024end}, BOFormer \cite{hung2025boformer} has recently explored pre-training policies via reinforcement learning on synthetic data in multi-objective context. However, to predict acquisition function values, its architecture remains heavily dependent on iterative GP modeling to generate posterior distributions. More recently, TAMO \cite{zhang2025context} further investigates fully amortized in-context optimization for multi-objective optimization by directly learning a query-generation policy. However, these amortized policy-learning methods both rely on hypervolume-oriented training objectives, whose computational cost can become substantial when scaled to massive pre-training regimes, thereby restricting the diversity of the pre-training corpus and hindering robust generalization to unseen complex problems.

\paragraph{Universal Regression Based on Transformers} Recently, Transformers \cite{vaswani2017attention} have become the preferred architecture for deep learning and foundation models. As model sizes and training data scaling, the models exhibit remarkable in-context learning (ICL) capabilities \cite{brown2020language}: they can make accurate predictions quickly in new tasks when prompted with only a few training examples, without any parameter update. Several research has demonstrated that Transformers can learn in-context simple logistic regression algorithms \cite{garg2022can,akyurek2022learning}. Beyond predicting labels, Transformer Neural Processes \cite{nguyen2022transformer} enhance the prediction with uncertainty as an alternative framework for uncertainty-aware meta learning. Prior-data Fitted Networks (PFNs) \cite{muller2021transformers} further show that Transformers can even learn to perform Bayesian inference with ICL. The developed universal regression models not only provide new insights into various model-based prediction topics \cite{hollmann2022tabpfn,hollmann2025accurate,dooley2023forecastpfn,song2024omnipred}, but also promote the regress-then-optimize framework towards a generalized optimization paradigm in both numerical \cite{muller2023pfns4bo,nguyen2023expt,rakotoarison2024context} and even linguistic space \cite{nguyen2024predicting,tan2025towards}. Building on this potential, we aim to develop a general optimization method for multi-objective scenarios, aligning more closely with real-world challenges that involve multi-criteria outputs.

\section{Methodology}
\label{sec:methodology}
\subsection{Framework}
\begin{figure}[!t]
\centering
\includegraphics[scale=0.41]{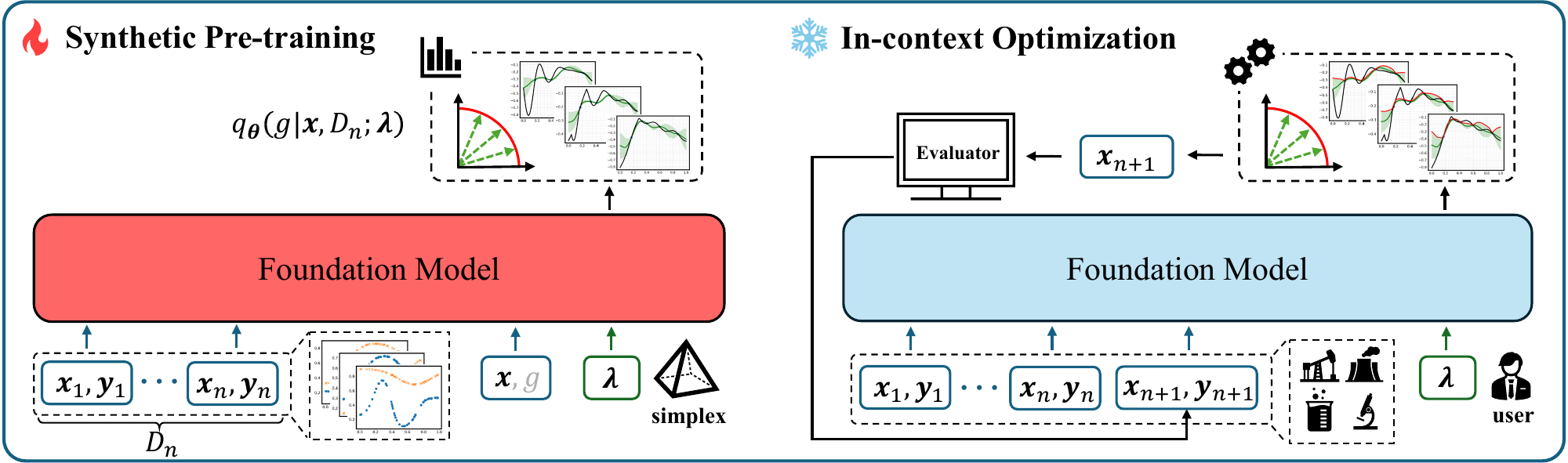}
\caption{Overview of FoMEMO framework, including synthetic pre-training stage (left) and in-context optimization stage (right).}
\label{fig:framework}
\end{figure}

Instead of building a separate surrogate model from scratch for every new problem encountered, the core concept of our method is to develop a universal foundation model. Our foundation model is pre-trained once only using a vast corpus of synthetic samples and preferences as context information, and outputs a large set of preference-wise aggregated posterior distributions for fast in-context optimization. We present the details of our framework in Figure \ref{fig:framework}, which includes a synthetic pre-training stage on the left and an in-context optimization stage on the right.

Specifically, in the synthetic pre-training stage, rather than accessing expensive real-world experiments, we implement a data sampling mechanism that allows us to heavily sample from it to generate a large set of synthetic data as the context inputs, including the trajectory pairs $D_n=\{(\boldsymbol{x}_i,\boldsymbol{y}_i)\}_{i=1}^n$, the query input $\boldsymbol{x}$ with the corresponding aggregation target $g=-s_{\boldsymbol{\lambda}}(\boldsymbol{x})$ masked, and the artificial preference $\boldsymbol{\lambda}$ sampled on the simplex, constructing a supervised learning problem for training the foundation model efficiently. We repeatedly synthesize training data as above and optimize the model's parameters $\boldsymbol{\theta}$ to predict the aggregated posterior distributions $q_{\boldsymbol{\theta}}(g|\boldsymbol{x},D_n;\boldsymbol{\lambda})$ conditioned on the contextual inputs. We will present the details of data generation in Section \ref{sec:data generation}.

Once the pre-training is completed, during the in-context optimization stage, users can tackle new problems by simply providing the evaluated trajectory from arbitrary unseen real-world applications, as well as the user's potential preferences (if any) as prompt inputs for the foundation model. Based on the aggregated posteriors predicted by the foundation model in a single forward pass, various acquisition functions can be easily derived to initiate fast in-context optimization to generate the next candidate for expensive evaluation, without the need for any additional model training or updates during the optimization process.

\subsection{Data Generation}
\label{sec:data generation}
While the pre-training distribution should ideally align with that of real-world functions, capturing such a distribution is often challenging due to limited empirical data. To bridge this gap, we implement a data generation mechanism designed to simulate a vast array of surrogate scenarios that the model may encounter. This approach enables the creation of a diverse yet computationally efficient dataset, bolstering the model's generalization capacity and enabling rapid adaptation to complex, unseen real-world optimization tasks.

Our data generation mechanism consists of three steps. Firstly, we sample general parameters for each set of the data, including the number of features $d$, the number of objectives $m$ and the length of trajectory $n$. Secondly, in order to simulate the multitude of implicit mappings that may exist in the real world, we sample a variety of synthetic functions from Gaussian processes (GPs). The choice of GPs is motivated by their cost-effectiveness and flexibility. Sampling from a GP prior is an easy and cost-effective method that enables the generation of a substantial number of latent mappings at a low cost. Besides, we can easily control the diversity of the generated functions by adjusting the kernel parameters. We employ a GP with the RBF kernel as it is a universal approximator \cite{micchelli2006universal} to encompass a wide range of unknown functions, which means in theory if trained long enough the foundation model will encounter synthetic functions that closely resemble true objective functions, enabling effective generalization. In the third step, we sample the observed trajectory $D_n=\{\boldsymbol{x}_{i},\boldsymbol{y}_{i}\}_{i=1}^{n}$ together with an arbitrary query pair $(\boldsymbol{x},\boldsymbol{y})$ with $d$-dimensional features and $m$-dimensional objectives from each sampled function. Since the aggregation target $g$ is determined only by $\boldsymbol{x}$ given $\boldsymbol{\lambda}$, after sampling artificial preferences from the simplex $\Delta=\{\boldsymbol{\lambda}\in\mathbb{R}_{+}^{m}\mid\|\boldsymbol{\lambda}\|_{1}=1\}$, we can calculate the aggregation targets efficiently for each query input given arbitrary preferences. We synthesize hundreds of millions of training data  using the aforementioned method, and mask the query aggregation targets to formulate a supervised prediction problem. This processing allows the foundation model to learn to predict the preference-wise aggregated posterior distribution $q_{\boldsymbol{\theta}}(g|\boldsymbol{x},D_n;\boldsymbol{\lambda})$ efficiently.

\subsection{Model Architecture}
Among the model architectures capable of learning uncertainty-calibrated predictions directly from data, we utilize the Prior-Data Fitted Networks (PFNs) \cite{muller2021transformers} as the base framework, as it provides a flexible paradigm for approximating the posterior distribution with arbitrary shape in a principled manner \cite{nagler2023statistical}. PFNs are based on a Transformer encoder that does not incorporate positional encodings, allowing the model to process data regardless of the order in which the input elements are presented. We use a variable-dimensional encoder to encode each input $\boldsymbol{x}$ and observation $\boldsymbol{y}$ into a token, allowing the model to adapt to cross-dimensional feature and objective spaces. An attention mask is implemented to enable each trajectory point can only attend to each other and the query point can only attend to the trajectory points. We also encode the sampled preference to inform the model to learn to predict the posterior distribution conditioned on which preference. To approximate the complex posterior distribution in high-dimensional feature and objective spaces, PFNs employ a piecewise constant distribution in the regression head. This approach converts the continuous distribution into a discrete distribution by utilizing multiple intervals, thereby transforming the regression task into a classification task to enhance the robustness during training. The detailed implementation can be referred to \cite{muller2021transformers}. To account for the distributional shifts in aggregated posteriors across varying objective dimensions, we propose an objective-aware regression head. This architecture dynamically routes the input to a specialized regressor tailored to the specific objective dimension of the problem at hand. By leveraging this dimension-specific mapping, the model can more accurately capture the differences of the aggregated posteriors with higher fidelity, thereby significantly enhancing its predictive precision in unseen landscapes across varying objective dimensions. Please refer to Appendices \ref{apd:Details of Foundation Model and Pre-training} and \ref{model_configuration_ablation} for detailed model configurations and the ablation analyses with different settings. 

\subsection{Training Details}
\label{sec:Training Details}
For any set of synthetic data generated from the sampling mechanism $p(D)$ presented in Section \ref{sec:data generation}, we optimize the following loss function, which is the cross-entropy between the held-out aggregation target and the model prediction:
\begin{equation}
\label{eq:loss function}
\mathbb{E}_{(\boldsymbol{x}, y)\cup D_n\sim p(D), \boldsymbol{\lambda}\sim\Delta, g=-s_{\boldsymbol{\lambda}}(\boldsymbol{x})}[-\log q_{\boldsymbol{\theta}}(g|\boldsymbol{x}, D_n;\boldsymbol{\lambda})].
\end{equation}
We continuously synthesize data through extensive sampling and optimize the model parameters to minimize the loss function, thereby progressively approaching the true posterior distribution $p(g|\boldsymbol{x},D_n;\boldsymbol{\lambda})$. 

Our final foundation model uses the PFNs architecture with 12 layers of Transformer, an embedding size of 512, a hidden size of 1024 in feed-forward layers, and 4-head attention, resulting in a total of 26.81M parameters. During training, we randomly sample each dataset from a large data space with feature dimensions ranging from 1 to 30, objective dimensions from 1 to 6, and trajectory lengths from 1 to 256. We sample a large variety of functions from GP. Specifically, we randomly sample the length scale kernel parameter $l\sim\Gamma(\alpha=3.0,\beta=6.0)$ for each feature and add a random observation noise $\epsilon\sim\mathcal N(0,10^{-4})$ to ensure function diversity. We do not sample the output scale parameter but instead set it to 1.0, as the predicted aggregation target will be normalized during inference. The final model is trained using 500 epochs and 2048 steps in each epoch with a batch size of 128 datasets, resulting more than 130 million datasets in total. The loss function is optimized using Adam optimizer \cite{kingma2014adam} with linear warmup and cosine annealing \cite{loshchilov2016sgdr} using a learning rate of $5\times10^{-5}$. The entire training process is conducted on 8 NVIDIA L20 GPUs and takes approximately 35 hours. We perform the pre-training only once and use the same model in all experiments in this paper.

\subsection{Generating Candidates}
Benefiting from the preference-wise aggregated posterior as the output basis, modeling the distribution of infinite sub-problems becomes possible. Acquisition functions using random scalarization and scalarization-based multi-objective indicators can be readily constructed. We have developed two types of acquisition functions including preference-based and preference-free variants in our work. 

\paragraph{Preference-based} We firstly derive the preference-based expected improvement (EI) and upper confidence bound (UCB) acquisition functions based on the single-objective scenario \cite{muller2023pfns4bo}:
\begin{equation}
\alpha_{\text{EI}}(\boldsymbol{x};\boldsymbol{\lambda})=\mathbb{E}_{q_{\boldsymbol{\theta}}(g|\boldsymbol{x}, D_n;\boldsymbol{\lambda})}[[g_{\boldsymbol{\lambda}}(\boldsymbol{x})-g_{\boldsymbol{\lambda}}^*]_{+}],
\end{equation}
\begin{equation}
\label{ucb}
\alpha_{\text{UCB}}(\boldsymbol{x};\boldsymbol{\lambda})={\mu}_{\boldsymbol{\lambda}}(\boldsymbol{x}) + \beta {\sigma}_{\boldsymbol{\lambda}}(\boldsymbol{x}),
\end{equation}
where $g_{\boldsymbol{\lambda}}^*$ is the best value of the aggregation $g_{\boldsymbol{\lambda}}(\boldsymbol{x})$ given an arbitrary preference $\boldsymbol{\lambda}$. ${\mu}_{\boldsymbol{\lambda}}(\boldsymbol{x})$ and ${\sigma}_{\boldsymbol{\lambda}}(\boldsymbol{x})$ are the mean and standard deviation of the aggregated posterior $q_{\boldsymbol{\theta}}(g|\boldsymbol{x}, D_n;\boldsymbol{\lambda})$, respectively. $\beta$ is a constant, which we set to 1.0 in this paper. As inference the acquisition functions is conditioned on the preference, during the optimization process, we randomly sample the preferences in each iteration and optimize the acquisition function to generate the next candidate solutions for evaluation.

\paragraph{Preference-free} To marginalize the preferences, we leverage a key theoretical insight of the popular used hypervolume metric: when employing an infinite set of reference vectors uniformly sampled from the unit hypersphere $S=\{\boldsymbol{w}\in\mathbb{R}_{+}^{m}\mid\|\boldsymbol{w}\|_{2}=1\}$, the hypervolume metric can be equivalently represented as the mean of a scalarization function \cite{deng2019approximating,zhang2020random}. Motivated by this, we firstly develop a preference-free acquisition function that utilizes the UCB utility defined in Equation (\ref{ucb}) to approximate the uncertainty-aware hypervolume improvement (UHVI) for any candidate $\boldsymbol{x}$, given the evaluated inputs $\boldsymbol{X}=\{\boldsymbol{x}_{i}\}_{i=1}^{n}$:
\begin{equation}
\begin{aligned}
\alpha_{\text{UHVI}}(\boldsymbol{x}) = c_m\mathbb{E}_{\boldsymbol{\lambda}\sim \Delta}[(c_{\boldsymbol{\lambda}})^m\max\{0, [\min_{\boldsymbol{x} \in \boldsymbol{X}}\{-g_{\boldsymbol{\lambda}}(\boldsymbol{x})\}]^m
-[-\alpha_{\text{UCB}}(\boldsymbol{x};\boldsymbol{\lambda})]^m\}],
\end{aligned}
\end{equation}
where $c_m=\frac{\pi^{m / 2}}{2^m \Gamma(m / 2+1)}$ is a constant that depends only on $m$. $c_{\boldsymbol{\lambda}}=\sqrt{\sum_{j=1}^{m} \frac{1}{\lambda_j^2}}$ is a transformation constant that depends only on $\boldsymbol{\lambda}$. The detailed derivations can be found in Appendix \ref{apd:UHVI}.

Analogous to the derivation of UHVI, we draw upon the R2 indicator \cite{hansen1994evaluating,brockhoff2012properties}, which is defined as the expected utility across a set of scalarization functions, to develop another preference-free acquisition function. This formulation, termed UR2I (uncertainty-aware R2 improvement), is designed to approximate the R2 improvement while explicitly accounting for predictive uncertainty:
\begin{equation}
\begin{aligned}
\alpha_{\text{UR2I}}(\boldsymbol{x}) = \mathbb{E}_{\boldsymbol{\lambda}\sim \Delta}[\max\{0, \min_{\boldsymbol{x} \in \boldsymbol{X}}\{-g_{\boldsymbol{\lambda}}(\boldsymbol{x})\}
-[-\alpha_{\text{UCB}}(\boldsymbol{x};\boldsymbol{\lambda})]\}].
\end{aligned}
\end{equation}

In terms of acquisition function selection, EI and UCB can guide candidate generation toward a user-specified trade-off or sample preferences across iterations when no preference is provided. In the absence of a specified preference, UHVI and UR2I offer additional options by aggregating improvements across multiple preferences within each iteration. We discuss these choices and provide principled and empirical guidance in Appendix~\ref{apd:acf_selection}.%
\section{Experiments}
\label{sec:experiments}

\subsection{Experimental Setting}
\label{sec:experimental_setting}

\paragraph{Baseline Algorithms} We compare our method against a wide range of classic and state-of-the-art MOBO algorithms. For consistency, MOEA/D-EGO \cite{zhang2009expensive}, TSEMO \cite{bradford2018efficient}, USEMO-EI \cite{belakaria2020uncertainty}, and DGEMO \cite{konakovic2020diversity} are all implemented using the DGEMO codebase based on Pymoo \cite{blank2020pymoo}. In addition, we evaluate qNEHVI \cite{daulton2021parallel}, qParEGO \cite{daulton2020differentiable}, and JES \cite{tu2022joint} using the BoTorch framework \cite{balandat2020botorch}. We further compare with Pareto set learning (PSL) \cite{lin2024smooth}, which learns mappings from preference vectors to Pareto-optimal solutions and also requires per-problem retraining. Finally, we include the recently proposed BOFormer \cite{hung2025boformer} and TAMO \cite{zhang2025context} as in-context baselines, which is most relevant to our work.

\paragraph{Benchmark Problems} We first evaluate the performance of all algorithms on 6 synthetic functions featuring a wide spectrum of landscapes, following the experimental setup in BOFormer \cite{hung2025boformer}. Subsequently, we conduct extensive experiments on 10 real-world engineering design problems from the RE benchmark suite \cite{tanabe2020easy}, which encompass diverse decision and objective space dimensions. Finally, we assess the methods on 4 hyperparameter optimization (HPO) tasks in a 3D object reconstruction problem \cite{mildenhall2021nerf}. Details of test problems are provided in Appendix \ref{apd:Benchmark Problems}.

\paragraph{Evaluation Protocol}
For all evaluated algorithms, each experiment is initialized with $2(d+1)$ randomly generated samples, followed by an optimization budget of 100 evaluations. To assess the quality of the Pareto front approximations produced by different algorithms, we employ the hypervolume (HV) indicator \cite{zitzler2002multiobjective} as the performance metric, given its strict monotonicity with respect to Pareto dominance. All optimization experiments are conducted over 10 independent runs, we report the mean performance alongside the standard deviation with 95\% confidence. Further technical details regarding the experimental configurations are provided in Appendix \ref{apd:Algorithm Hyperparameters} and \ref{apd:Performance Metrics}.

\subsection{Main Results}
\label{sec:main_results}

\begin{table}[!t]
\centering
\caption{Means (stds) of hypervolume metrics ($\uparrow$) on partial problems over 10 independent runs. The best, second-best, and third-best mean results for each problem are highlighted with dark gray, gray, and light gray backgrounds, respectively.}
\label{tab:selected_exp}
\renewcommand{\arraystretch}{1.5}
\scalebox{0.62}{
\begin{tabular}{ccccccc}
\toprule[1pt]
Algorithm & BC & ABD & Four Bar Truss & Gear Train & Materials & Mic \\
\hline
MOEA/D-EGO & 7.015e-01(7.5e-03) & 6.934e-01(1.1e-02) & 7.462e-01(1.4e-02) & 5.623e-01(1.3e-01) & 1.166e+00(2.2e-03) & 1.257e+00(4.9e-03) \\
TSEMO & 8.056e-01(9.1e-04) & 7.114e-01(9.4e-03) & 8.506e-01(4.7e-04) & 8.092e-01(2.3e-02) & 1.160e+00(5.1e-03) & 1.251e+00(4.8e-03) \\
USEMO-EI & 7.750e-01(1.2e-02) & 7.298e-01(1.1e-02) & 8.232e-01(4.3e-03) & 6.433e-01(4.0e-02) & 1.167e+00(2.8e-03) & 1.255e+00(5.4e-03) \\
DGEMO & \cellcolor{gray!50}\textbf{8.087e-01(1.3e-03)} & 5.918e-01(6.8e-02) & 8.526e-01(9.8e-04) & 8.859e-01(3.2e-02) & 1.172e+00(3.2e-03) & \cellcolor{gray!30}\textbf{1.273e+00(2.1e-02)} \\
qNEHVI & 7.962e-01(1.7e-03) & 7.181e-01(1.0e-02) & 8.244e-01(3.8e-03) & 8.387e-01(2.6e-02) & 1.144e+00(1.1e-02) & 1.198e+00(1.9e-02) \\
qParEGO & 7.843e-01(4.2e-03) & 7.032e-01(9.1e-03) & 7.917e-01(1.7e-02) & 7.813e-01(3.1e-02) & 1.150e+00(3.3e-03) & 1.191e+00(2.0e-02) \\
JES & 7.765e-01(3.1e-03) & 6.598e-01(1.3e-02) & 7.918e-01(1.0e-02) & 6.096e-01(4.5e-02) & 1.156e+00(3.2e-03) & 1.212e+00(1.3e-02) \\
PSL & 6.708e-01(1.8e-02) & 4.548e-01(8.9e-02) & 8.655e-01(2.8e-03) & 8.488e-01(2.2e-02) & 1.073e+00(4.5e-02) & 1.123e+00(2.2e-02) \\
BOFormer & 7.187e-01(1.4e-02) & 6.383e-01(2.5e-02) & 7.308e-01(1.4e-02) & 3.818e-01(5.7e-02) & 1.161e+00(3.6e-03) & 1.244e+00(6.3e-03) \\
TAMO & 7.411e-01(8.7e-03) & 7.270e-01(4.3e-03) & 8.069e-01(3.9e-03) & 6.460e-01(8.4e-05) & 1.164e+00(2.7e-03) & 1.244e+00(1.9e-02) \\
\hline
FoMEMO-EI & \cellcolor{gray!10}\textbf{8.066e-01(6.2e-04)} & \cellcolor{gray!10}\textbf{7.308e-01(7.9e-03)} & \cellcolor{gray!10}\textbf{8.667e-01(2.0e-03)} & \cellcolor{gray!30}\textbf{9.107e-01(8.1e-03)} & 1.173e+00(2.0e-03) & 1.261e+00(2.1e-02) \\
FoMEMO-UCB & \cellcolor{gray!30}\textbf{8.075e-01(5.8e-04)} & \cellcolor{gray!30}\textbf{7.369e-01(8.5e-03)} & \cellcolor{gray!30}\textbf{8.722e-01(1.2e-03)} & \cellcolor{gray!50}\textbf{9.270e-01(4.5e-03)} & \cellcolor{gray!10}\textbf{1.175e+00(1.9e-03)} & 1.262e+00(1.8e-02) \\
FoMEMO-UHVI & 8.052e-01(4.9e-04) & \cellcolor{gray!50}\textbf{7.430e-01(8.4e-03)} & 8.609e-01(1.3e-03) & 8.573e-01(9.6e-03) & \cellcolor{gray!30}\textbf{1.176e+00(2.4e-03)} & \cellcolor{gray!10}\textbf{1.271e+00(1.6e-02)} \\
FoMEMO-UR2I & 8.016e-01(1.7e-03) & 7.126e-01(9.1e-03) & \cellcolor{gray!50}\textbf{8.741e-01(5.0e-04)} & \cellcolor{gray!10}\textbf{8.961e-01(1.1e-02)} & \cellcolor{gray!50}\textbf{1.179e+00(2.2e-03)} & \cellcolor{gray!50}\textbf{1.274e+00(1.2e-02)} \\
\bottomrule
\end{tabular}}
\end{table}

Table \ref{tab:selected_exp} summarizes the sequential optimization results across partial problems covering the three benchmark families. The complete performance results are provided in Tables \ref{tab:bbob exp}, \ref{tab:re exp}, and \ref{tab:hpo exp} in Appendix \ref{apd:complete performance results}, the query time analysis is shown in Appendix \ref{apd: analysis of runtime}. The experimental results demonstrate that our framework achieves strong performance with favorable query efficiency in the majority of cases. This underscores that by pre-training the foundation model on an extensive and diverse corpus of synthetic data, the model showcases robust adaptability and generalization for problems that it has not encountered during training.

\section{Further Discussions}
\label{sec:further_discussions}

\subsection{Benefits of Aggregated Posterior Learning}
\label{sec:benefits_aggregated_posterior_learning}
Our main contribution is a scalable, decomposition-based pre-training paradigm that reformulates MOBO as preference-wise aggregated posterior learning, which is motivated by the need for broad and efficient adaptation across diverse emerging real-world applications. In Appendix~\ref{app:aggregated_posterior_learning}, we provide a detailed analysis of the benefits of aggregated posterior learning as the central mechanism underlying our paradigm. We first examine the bottlenecks of existing MOBO approaches including repeated surrogate modeling, posterior fidelity limitations, acquisition optimization overhead and training scalability challenge. We then explain how aggregated posterior learning serves as the key component that supports our paradigm and addresses these bottlenecks by simultaneously enabling amortized surrogate modeling, high-fidelity posterior prediction, efficient acquisition optimization and scalable pre-training. Finally, we summarize the ablation evidence to identify the sources of the observed performance gains and wall-time reductions, including posterior calibration and downstream optimization comparisons with GP- and PFN-based baselines (Appendix~\ref{pfn_based_methods_ablation}), training-data generation efficiency (discussed below in Section~~\ref{sec:scalable_training_advantage}) and query-time advantages over conventional MOBO and PFN-based methods (Tables~\ref{tab:query time} and~\ref{tab:pfns4bo-query-time}).

\subsection{Scalable Training Advantage}
\label{sec:scalable_training_advantage}
We report the mean runtime of metric computation during training using hypervolume and Tchebycheff aggregation to emphasize the advantage of our scalable training. As shown in Table \ref{tab:metric computation time}, although neither hypervolume nor Tchebycheff is particularly costly in low-dimensional objective spaces, our main point is that achieving strong generalization typically requires pre-training on massive synthetic data (130 million datasets in our work) to cover diverse dimensions, trajectory lengths, and function properties. At this scale, even modest per-instance cost differences become substantial. As shown in the results, even for a single dataset with a sample length of 128 and low objective dimensions, the time of computing hypervolume metric is already orders of magnitude slower than Tchebycheff, and its runtime increases rapidly with the number of objectives, whereas Tchebycheff remains much more efficient. This makes our framework better suited for large-scale training data generation and thus better aligned with the foundation model paradigm.

\begin{table}[htbp]
\centering
\caption{Mean runtime of metric computation on a single dataset with length of 128 over 10 independent runs.}
\label{tab:metric computation time}
\renewcommand{\arraystretch}{1.3}
\scalebox{0.78}{
\begin{tabular}{lcccc}
\toprule
Problem & \# Objectives & Tchebycheff (s) & Hypervolume (s) & Hypervolume / Tchebycheff \\
\midrule
Four Bar Truss & 2 & 6.10e-05 & 4.14e-04 & 6.79 \\
Speed Reducer & 3 & 6.45e-05 & 2.55e-03 & 39.53 \\
Car Side Impact & 4 & 6.62e-05 & 1.23e-01 & 1853.56 \\
Water Resource Planning & 6 & 8.56e-05 & 3.97e+01 & 463594.84 \\
\bottomrule
\end{tabular}}
\end{table}

\subsection{Predictive Accuracy During Pre-training}
\begin{figure}[htbp]
\centering
\includegraphics[scale=0.23]{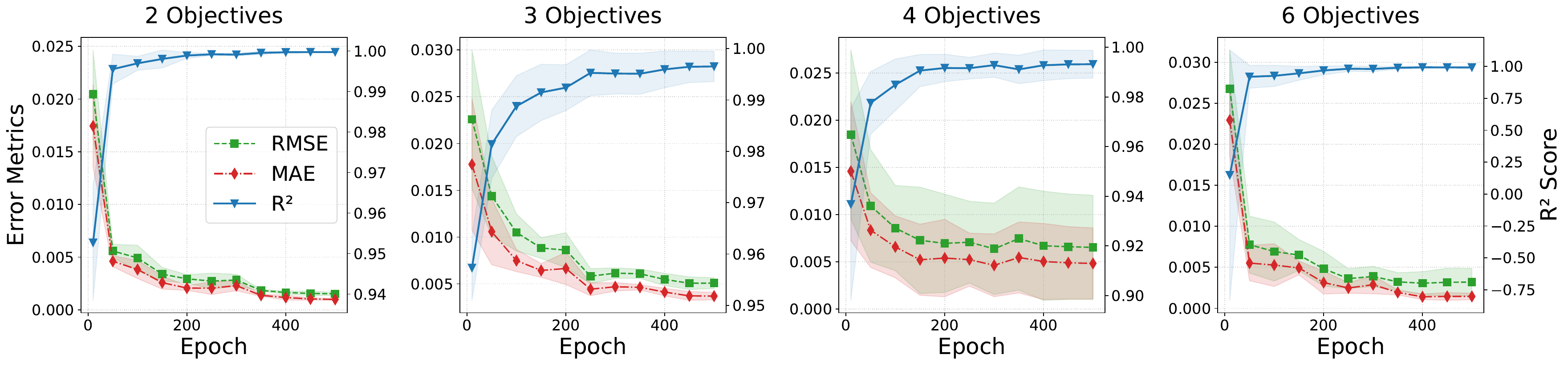}
\caption{The evolution of modeling accuracy in terms of regression error metrics (RMSE, MAE) and $R^2$ score during the training process.}
\label{fig:regression metric checkpoint}
\end{figure}

Since the efficacy of our method relies on modeling fidelity, in addition to providing a posterior visualization for a simple problem (see Appendix~\ref{apd:Visualization of Aggregated Posteriors}), we further analyze the model's ability to fit aggregation targets across unseen problems with varying objective dimensions during pre-training. Specifically, we monitor the accuracy of the predicted means throughout the training process, as shown in Figure~\ref{fig:regression metric checkpoint}. We randomly sample 1000 datasets from four engineering design problems and partition them into training and testing sets using a 7:3 ratio. For each of the five independent experimental trials, a preference vector is randomly sampled to generate the aggregation targets. We employe RMSE and MAE metrics to quantify point-wise accuracy, alongside the $R^2$ score to evaluate the model's capacity to capture the underlying data trends. We observe that as training progresses, the model’s predictive accuracy on unseen problems improves consistently. This evolution in modeling fidelity provides a fundamental explanation for the superior optimization performance demonstrated by our method.

\subsection{Scalability Analysis}

\paragraph{Scalability Within Pre-training Dimensions}We first investigate the model's scalability to higher objective and decision dimensions within its pre-training ranges, with detailed results and analysis provided in Appendix \ref{app:scalability_within_pretraining_dimensions}. FoMEMO demonstrates favorable scalability across both objective and decision dimensions. On many-objective engineering problems, it maintains robust performance by directly modeling preference-wise aggregated posteriors, thereby avoiding costly hypervolume partitioning and reducing scalarization-induced approximation errors. Results on 30-dimensional WFG4 suggest that FoMEMO remains effective at the boundary of its pre-training decision dimensionality.

\paragraph{Scalability Beyond Pre-training Dimensions} We further examine scalability beyond the pre-training ranges in decision dimensionality, objective dimensionality, and observation count using 40-dimensional WFG4, 9-objective Car Cab Design, and Gear Train Design with 310 evaluations. FoMEMO achieves the best final HV on WFG4 and Gear Train Design and competitive IGD+ on Car Cab Design. Its average query time is substantially lower on the 40-dimensional and 9-objective tests and remains manageable in the 310-evaluation setting. Detailed experimental setup and results are provided in Appendix~\ref{app:scalability_beyond_pretraining_dimensions}.

\paragraph{Scalability in Parallel Optimization}In addition to sequential optimization, we investigate the scalability of our method in the context of parallel optimization. For preference-based acquisition functions, we generate a batch of $q$ candidates by conditioning $q$ random preferences sampled in each iteration. For preference-free variants, since approximating HV or R2 improvement via a finite set of random preferences introduces inherent stochastic deviation, we leverage this characteristic by independently repeating UHVI or UR2I calculations $q$ times to produce $q$ candidates for evaluation. We evaluate this parallel scheme under batch sizes of $q=5$ and $q=10$. Our results, detailed in Appendix \ref{app:scalability_in_parallel_optimization}, demonstrate that even with these straightforward parallelization strategies, our preference-based methods consistently outperform MOBO algorithms that rely on well-designed batch selection heuristics. This performance gap highlights the advantage of using a foundation model pre-trained on a large variety of synthetic landscapes for optimization. Moreover, it demonstrates that modeling the preference-wise aggregated posterior is inherently well-suited for parallelization, providing a more robust and natural framework for multi-objective optimization without sophisticated batch selection processes.

\subsection{Method Applicability}
We conduct a two-step exploration to investigate the applicable scope of our methodology and discuss potential extensions in Appendix \ref{apd:method applicability}. The mixed-kernel experiments show that increasing the diversity of training kernels can be beneficial, but does not uniformly improve performance across all tasks and acquisition functions. In addition, the Weierstrass-Takagi test illustrates that when the target objectives exhibit function regularity that differs substantially from the training prior, prior mismatch can affect both posterior prediction and downstream optimization performance. These results suggest that prior diversity is useful, but the alignment between the training prior and the target function characteristics plays a more important role in determining generalization.

\section{Conclusion and Future Work}
\label{sec:conclusion}
We propose a novel paradigm named FoMEMO, which establishes a foundation model to address a wide range of expensive multi-objective optimization problems in an in-context manner. 
Extensive experiments verify that our method exhibits remarkable adaptability and generalization in various synthetic problems and real-world applications, showing strong optimization performance and favorable query efficiency compared to existing methods. Future work includes incorporating domain-specific priors, adaptive prior enrichment, or lightweight fine-tuning to improve robustness when target problems deviate from the synthetic pre-training distribution. We hope this work provides a foundation for future research on scalable, generalizable, and domain-adaptive optimization models for expensive multi-objective optimization.

\section*{Acknowledgments}
The work described in this paper was supported by the Research Grants Council of the Hong Kong Special Administrative Region, China (GRF Project No. CityU 11212524) and the National Natural Science Foundation of China (Project No. 62276223).
%




\bibliographystyle{plainnat}
\bibliography{references}

@article{xue2024offline,
  title={Offline multi-objective optimization},
  author={Xue, Ke and Tan, Rong-Xi and Huang, Xiaobin and Qian, Chao},
  journal={arXiv preprint arXiv:2406.03722},
  year={2024}
}

@article{yuan2024paretoflow,
  title={ParetoFlow: Guided Flows in Multi-Objective Optimization},
  author={Yuan, Ye and Chen, Can and Pal, Christopher and Liu, Xue},
  journal={arXiv preprint arXiv:2412.03718},
  year={2024}
}

@inproceedings{ying2019bench,
  title={Nas-bench-101: Towards reproducible neural architecture search},
  author={Ying, Chris and Klein, Aaron and Christiansen, Eric and Real, Esteban and Murphy, Kevin and Hutter, Frank},
  booktitle={International conference on machine learning},
  pages={7105--7114},
  year={2019},
  organization={PMLR}
}

@article{tanabe2020easy,
  title={An easy-to-use real-world multi-objective optimization problem suite},
  author={Tanabe, Ryoji and Ishibuchi, Hisao},
  journal={Applied Soft Computing},
  volume={89},
  pages={106078},
  year={2020},
  publisher={Elsevier}
}

@article{dara2022machine,
  title={Machine learning in drug discovery: a review},
  author={Dara, Suresh and Dhamercherla, Swetha and Jadav, Surender Singh and Babu, CH Madhu and Ahsan, Mohamed Jawed},
  journal={Artificial intelligence review},
  volume={55},
  number={3},
  pages={1947--1999},
  year={2022},
  publisher={Springer}
}

@article{kingma2014adam,
  title={Adam: A method for stochastic optimization},
  author={Kingma, Diederik P and Ba, Jimmy},
  journal={arXiv preprint arXiv:1412.6980},
  year={2014}
}

@article{loshchilov2016sgdr,
  title={Sgdr: Stochastic gradient descent with warm restarts},
  author={Loshchilov, Ilya and Hutter, Frank},
  journal={arXiv preprint arXiv:1608.03983},
  year={2016}
}

@inproceedings{muller2023pfns4bo,
  title={Pfns4bo: In-context learning for bayesian optimization},
  author={M{\"u}ller, Samuel and Feurer, Matthias and Hollmann, Noah and Hutter, Frank},
  booktitle={International Conference on Machine Learning},
  pages={25444--25470},
  year={2023},
  organization={PMLR}
}

@article{deng2019approximating,
  title={Approximating hypervolume and hypervolume contributions using polar coordinate},
  author={Deng, Jingda and Zhang, Qingfu},
  journal={IEEE Transactions on Evolutionary Computation},
  volume={23},
  number={5},
  pages={913--918},
  year={2019},
  publisher={IEEE}
}

@inproceedings{zhang2020random,
  title={Random hypervolume scalarizations for provable multi-objective black box optimization},
  author={Zhang, Richard and Golovin, Daniel},
  booktitle={International conference on machine learning},
  pages={11096--11105},
  year={2020},
  organization={PMLR}
}

@article{zhang2009expensive,
  title={Expensive multiobjective optimization by MOEA/D with Gaussian process model},
  author={Zhang, Qingfu and Liu, Wudong and Tsang, Edward and Virginas, Botond},
  journal={IEEE Transactions on Evolutionary Computation},
  volume={14},
  number={3},
  pages={456--474},
  year={2009},
  publisher={IEEE}
}

@article{bradford2018efficient,
  title={Efficient multiobjective optimization employing Gaussian processes, spectral sampling and a genetic algorithm},
  author={Bradford, Eric and Schweidtmann, Artur M and Lapkin, Alexei},
  journal={Journal of global optimization},
  volume={71},
  number={2},
  pages={407--438},
  year={2018},
  publisher={Springer}
}

@inproceedings{belakaria2020uncertainty,
  title={Uncertainty-aware search framework for multi-objective Bayesian optimization},
  author={Belakaria, Syrine and Deshwal, Aryan and Jayakodi, Nitthilan Kannappan and Doppa, Janardhan Rao},
  booktitle={Proceedings of the AAAI Conference on Artificial Intelligence},
  volume={34},
  number={06},
  pages={10044--10052},
  year={2020}
}

@article{konakovic2020diversity,
  title={Diversity-guided multi-objective bayesian optimization with batch evaluations},
  author={Konakovic Lukovic, Mina and Tian, Yunsheng and Matusik, Wojciech},
  journal={Advances in Neural Information Processing Systems},
  volume={33},
  pages={17708--17720},
  year={2020}
}

@article{daulton2020differentiable,
  title={Differentiable expected hypervolume improvement for parallel multi-objective Bayesian optimization},
  author={Daulton, Samuel and Balandat, Maximilian and Bakshy, Eytan},
  journal={Advances in neural information processing systems},
  volume={33},
  pages={9851--9864},
  year={2020}
}

@article{daulton2021parallel,
  title={Parallel bayesian optimization of multiple noisy objectives with expected hypervolume improvement},
  author={Daulton, Samuel and Balandat, Maximilian and Bakshy, Eytan},
  journal={Advances in neural information processing systems},
  volume={34},
  pages={2187--2200},
  year={2021}
}

@article{balandat2020botorch,
  title={BoTorch: A framework for efficient Monte-Carlo Bayesian optimization},
  author={Balandat, Maximilian and Karrer, Brian and Jiang, Daniel and Daulton, Samuel and Letham, Ben and Wilson, Andrew G and Bakshy, Eytan},
  journal={Advances in neural information processing systems},
  volume={33},
  pages={21524--21538},
  year={2020}
}

@article{knowles2006parego,
  title={ParEGO: A hybrid algorithm with on-line landscape approximation for expensive multiobjective optimization problems},
  author={Knowles, Joshua},
  journal={IEEE transactions on evolutionary computation},
  volume={10},
  number={1},
  pages={50--66},
  year={2006},
  publisher={IEEE}
}

@book{hansen1994evaluating,
  title={Evaluating the quality of approximations to the non-dominated set},
  author={Hansen, Michael Pilegaard and Jaszkiewicz, Andrzej},
  year={1994},
  publisher={IMM, Department of Mathematical Modelling, Technical Universityof Denmark}
}

@inproceedings{brockhoff2012properties,
  title={On the properties of the R2 indicator},
  author={Brockhoff, Dimo and Wagner, Tobias and Trautmann, Heike},
  booktitle={Proceedings of the 14th annual conference on Genetic and evolutionary computation},
  pages={465--472},
  year={2012}
}

@article{williams1995gaussian,
  title={Gaussian processes for regression},
  author={Williams, Christopher and Rasmussen, Carl},
  journal={Advances in neural information processing systems},
  volume={8},
  year={1995}
}

@book{williams2006gaussian,
  title={Gaussian processes for machine learning},
  author={Williams, Christopher KI and Rasmussen, Carl Edward},
  volume={2},
  year={2006},
  publisher={MIT press Cambridge, MA}
}

@article{frazier2018tutorial,
  title={A tutorial on Bayesian optimization},
  author={Frazier, Peter I},
  journal={arXiv preprint arXiv:1807.02811},
  year={2018}
}

@article{shahriari2015taking,
  title={Taking the human out of the loop: A review of Bayesian optimization},
  author={Shahriari, Bobak and Swersky, Kevin and Wang, Ziyu and Adams, Ryan P and De Freitas, Nando},
  journal={Proceedings of the IEEE},
  volume={104},
  number={1},
  pages={148--175},
  year={2015},
  publisher={IEEE}
}

@book{garnett2023bayesian,
  title={Bayesian optimization},
  author={Garnett, Roman},
  year={2023},
  publisher={Cambridge University Press}
}

@article{zitzler2002multiobjective,
  title={Multiobjective evolutionary algorithms: a comparative case study and the strength Pareto approach},
  author={Zitzler, Eckart and Thiele, Lothar},
  journal={IEEE transactions on Evolutionary Computation},
  volume={3},
  number={4},
  pages={257--271},
  year={2002},
  publisher={IEEE}
}

@article{choo1983proper,
  title={Proper efficiency in nonconvex multicriteria programming},
  author={Choo, Eng Ung and Atkins, Derek R},
  journal={Mathematics of Operations Research},
  volume={8},
  number={3},
  pages={467--470},
  year={1983},
  publisher={INFORMS}
}

@inproceedings{paria2020flexible,
  title={A flexible framework for multi-objective bayesian optimization using random scalarizations},
  author={Paria, Biswajit and Kandasamy, Kirthevasan and P{\'o}czos, Barnab{\'a}s},
  booktitle={Uncertainty in Artificial Intelligence},
  pages={766--776},
  year={2020},
  organization={PMLR}
}

@article{emmerich2006single,
  title={Single-and multiobjective evolutionary optimization assisted by Gaussian random field metamodels},
  author={Emmerich, Michael TM and Giannakoglou, Kyriakos C and Naujoks, Boris},
  journal={IEEE Transactions on Evolutionary Computation},
  volume={10},
  number={4},
  pages={421--439},
  year={2006},
  publisher={IEEE}
}

@article{jones1998efficient,
  title={Efficient global optimization of expensive black-box functions},
  author={Jones, Donald R and Schonlau, Matthias and Welch, William J},
  journal={Journal of Global optimization},
  volume={13},
  number={4},
  pages={455--492},
  year={1998},
  publisher={Springer}
}

@article{wada2019bayesian,
  title={Bayesian optimization for multi-objective optimization and multi-point search},
  author={Wada, Takashi and Hino, Hideitsu},
  journal={arXiv preprint arXiv:1905.02370},
  year={2019}
}

@article{vaswani2017attention,
  title={Attention is all you need},
  author={Vaswani, Ashish and Shazeer, Noam and Parmar, Niki and Uszkoreit, Jakob and Jones, Llion and Gomez, Aidan N and Kaiser, {\L}ukasz and Polosukhin, Illia},
  journal={Advances in neural information processing systems},
  volume={30},
  year={2017}
}

@article{brown2020language,
  title={Language models are few-shot learners},
  author={Brown, Tom and Mann, Benjamin and Ryder, Nick and Subbiah, Melanie and Kaplan, Jared D and Dhariwal, Prafulla and Neelakantan, Arvind and Shyam, Pranav and Sastry, Girish and Askell, Amanda and others},
  journal={Advances in neural information processing systems},
  volume={33},
  pages={1877--1901},
  year={2020}
}

@article{garg2022can,
  title={What can transformers learn in-context? a case study of simple function classes},
  author={Garg, Shivam and Tsipras, Dimitris and Liang, Percy S and Valiant, Gregory},
  journal={Advances in neural information processing systems},
  volume={35},
  pages={30583--30598},
  year={2022}
}

@article{akyurek2022learning,
  title={What learning algorithm is in-context learning? investigations with linear models},
  author={Aky{\"u}rek, Ekin and Schuurmans, Dale and Andreas, Jacob and Ma, Tengyu and Zhou, Denny},
  journal={arXiv preprint arXiv:2211.15661},
  year={2022}
}

@article{nguyen2022transformer,
  title={Transformer neural processes: Uncertainty-aware meta learning via sequence modeling},
  author={Nguyen, Tung and Grover, Aditya},
  journal={arXiv preprint arXiv:2207.04179},
  year={2022}
}

@article{muller2021transformers,
  title={Transformers can do bayesian inference},
  author={M{\"u}ller, Samuel and Hollmann, Noah and Arango, Sebastian Pineda and Grabocka, Josif and Hutter, Frank},
  journal={arXiv preprint arXiv:2112.10510},
  year={2021}
}

@article{hollmann2022tabpfn,
  title={Tabpfn: A transformer that solves small tabular classification problems in a second},
  author={Hollmann, Noah and M{\"u}ller, Samuel and Eggensperger, Katharina and Hutter, Frank},
  journal={arXiv preprint arXiv:2207.01848},
  year={2022}
}

@article{hollmann2025accurate,
  title={Accurate predictions on small data with a tabular foundation model},
  author={Hollmann, Noah and M{\"u}ller, Samuel and Purucker, Lennart and Krishnakumar, Arjun and K{\"o}rfer, Max and Hoo, Shi Bin and Schirrmeister, Robin Tibor and Hutter, Frank},
  journal={Nature},
  volume={637},
  number={8045},
  pages={319--326},
  year={2025},
  publisher={Nature Publishing Group UK London}
}

@article{dooley2023forecastpfn,
  title={Forecastpfn: Synthetically-trained zero-shot forecasting},
  author={Dooley, Samuel and Khurana, Gurnoor Singh and Mohapatra, Chirag and Naidu, Siddartha V and White, Colin},
  journal={Advances in Neural Information Processing Systems},
  volume={36},
  pages={2403--2426},
  year={2023}
}

@article{song2024omnipred,
  title={Omnipred: Language models as universal regressors},
  author={Song, Xingyou and Li, Oscar and Lee, Chansoo and Yang, Bangding and Peng, Daiyi and Perel, Sagi and Chen, Yutian},
  journal={arXiv preprint arXiv:2402.14547},
  year={2024}
}

@article{rakotoarison2024context,
  title={In-context freeze-thaw bayesian optimization for hyperparameter optimization},
  author={Rakotoarison, Herilalaina and Adriaensen, Steven and Mallik, Neeratyoy and Garibov, Samir and Bergman, Edward and Hutter, Frank},
  journal={arXiv preprint arXiv:2404.16795},
  year={2024}
}

@article{nguyen2023expt,
  title={Expt: Synthetic pretraining for few-shot experimental design},
  author={Nguyen, Tung and Agrawal, Sudhanshu and Grover, Aditya},
  journal={Advances in Neural Information Processing Systems},
  volume={36},
  pages={45856--45869},
  year={2023}
}

@article{nguyen2024predicting,
  title={Predicting from Strings: Language Model Embeddings for Bayesian Optimization},
  author={Nguyen, Tung and Zhang, Qiuyi and Yang, Bangding and Lee, Chansoo and Bornschein, Jorg and Miao, Yingjie and Perel, Sagi and Chen, Yutian and Song, Xingyou},
  journal={arXiv preprint arXiv:2410.10190},
  year={2024}
}

@article{tan2025towards,
  title={Towards Universal Offline Black-Box Optimization via Learning Language Model Embeddings},
  author={Tan, Rong-Xi and Chen, Ming and Xue, Ke and Wang, Yao and Wang, Yaoyuan and Fu, Sheng and Qian, Chao},
  journal={arXiv preprint arXiv:2506.07109},
  year={2025}
}

@inproceedings{trabucco2021conservative,
  title={Conservative objective models for effective offline model-based optimization},
  author={Trabucco, Brandon and Kumar, Aviral and Geng, Xinyang and Levine, Sergey},
  booktitle={International Conference on Machine Learning},
  pages={10358--10368},
  year={2021},
  organization={PMLR}
}

@article{yu2021roma,
  title={Roma: Robust model adaptation for offline model-based optimization},
  author={Yu, Sihyun and Ahn, Sungsoo and Song, Le and Shin, Jinwoo},
  journal={Advances in Neural Information Processing Systems},
  volume={34},
  pages={4619--4631},
  year={2021}
}

@article{chen2023parallel,
  title={Parallel-mentoring for offline model-based optimization},
  author={Chen, Can Sam and Beckham, Christopher and Liu, Zixuan and Liu, Xue Steve and Pal, Chris},
  journal={Advances in Neural Information Processing Systems},
  volume={36},
  pages={76619--76636},
  year={2023}
}

@inproceedings{nagler2023statistical,
  title={Statistical foundations of prior-data fitted networks},
  author={Nagler, Thomas},
  booktitle={International Conference on Machine Learning},
  pages={25660--25676},
  year={2023},
  organization={PMLR}
}

@inproceedings{daulton2022robust,
  title={Robust multi-objective bayesian optimization under input noise},
  author={Daulton, Samuel and Cakmak, Sait and Balandat, Maximilian and Osborne, Michael A and Zhou, Enlu and Bakshy, Eytan},
  booktitle={International Conference on Machine Learning},
  pages={4831--4866},
  year={2022},
  organization={PMLR}
}

@article{zhang2023hypervolume,
  title={Hypervolume maximization: A geometric view of pareto set learning},
  author={Zhang, Xiaoyuan and Lin, Xi and Xue, Bo and Chen, Yifan and Zhang, Qingfu},
  journal={Advances in Neural Information Processing Systems},
  volume={36},
  pages={38902--38929},
  year={2023}
}

@article{song2024vizier,
  title={The vizier gaussian process bandit algorithm},
  author={Song, Xingyou and Zhang, Qiuyi and Lee, Chansoo and Fertig, Emily and Huang, Tzu-Kuo and Belenki, Lior and Kochanski, Greg and Ariafar, Setareh and Vasudevan, Srinivas and Perel, Sagi and others},
  journal={arXiv preprint arXiv:2408.11527},
  year={2024}
}

@article{mildenhall2021nerf,
  title={Nerf: Representing scenes as neural radiance fields for view synthesis},
  author={Mildenhall, Ben and Srinivasan, Pratul P and Tancik, Matthew and Barron, Jonathan T and Ramamoorthi, Ravi and Ng, Ren},
  journal={Communications of the ACM},
  volume={65},
  number={1},
  pages={99--106},
  year={2021},
  publisher={ACM New York, NY, USA}
}

@article{hung2025boformer,
  title={Boformer: Learning to solve multi-objective bayesian optimization via non-markovian rl},
  author={Hung, Yu-Heng and Lin, Kai-Jie and Lin, Yu-Heng and Wang, Chien-Yi and Sun, Cheng and Hsieh, Ping-Chun},
  journal={arXiv preprint arXiv:2505.21974},
  year={2025}
}

@article{kerbl20233d,
  title={3D Gaussian splatting for real-time radiance field rendering.},
  author={Kerbl, Bernhard and Kopanas, Georgios and Leimk{\"u}hler, Thomas and Drettakis, George},
  journal={ACM Trans. Graph.},
  volume={42},
  number={4},
  pages={139--1},
  year={2023}
}

@article{lin2024smooth,
  title={Smooth tchebycheff scalarization for multi-objective optimization},
  author={Lin, Xi and Zhang, Xiaoyuan and Yang, Zhiyuan and Liu, Fei and Wang, Zhenkun and Zhang, Qingfu},
  journal={arXiv preprint arXiv:2402.19078},
  year={2024}
}

@article{huband2006review,
  title={A review of multiobjective test problems and a scalable test problem toolkit},
  author={Huband, Simon and Hingston, Philip and Barone, Luigi and While, Lyndon},
  journal={IEEE Transactions on Evolutionary Computation},
  volume={10},
  number={5},
  pages={477--506},
  year={2006},
  publisher={IEEE}
}

@article{zhao2024many,
  title={Many-to-Few Decomposition: Linking R2-Based and Decomposition-Based Multiobjective Efficient Global Optimization Algorithms},
  author={Zhao, Liang and Huang, Xiaobin and Qian, Chao and Zhang, Qingfu},
  journal={IEEE Transactions on Evolutionary Computation},
  year={2024},
  publisher={IEEE}
}

@article{blank2020pymoo,
  title={Pymoo: Multi-objective optimization in python},
  author={Blank, Julian and Deb, Kalyanmoy},
  journal={Ieee access},
  volume={8},
  pages={89497--89509},
  year={2020},
  publisher={IEEE}
}

@article{tu2022joint,
  title={Joint entropy search for multi-objective Bayesian optimization},
  author={Tu, Ben and Gandy, Axel and Kantas, Nikolas and Shafei, Behrang},
  journal={Advances in Neural Information Processing Systems},
  volume={35},
  pages={9922--9938},
  year={2022}
}

@article{hvarfner2022joint,
  title={Joint entropy search for maximally-informed Bayesian optimization},
  author={Hvarfner, Carl and Hutter, Frank and Nardi, Luigi},
  journal={Advances in Neural Information Processing Systems},
  volume={35},
  pages={11494--11506},
  year={2022}
}

@article{ament2023unexpected,
  title={Unexpected improvements to expected improvement for bayesian optimization},
  author={Ament, Sebastian and Daulton, Samuel and Eriksson, David and Balandat, Maximilian and Bakshy, Eytan},
  journal={Advances in Neural Information Processing Systems},
  volume={36},
  pages={20577--20612},
  year={2023}
}

@article{micchelli2006universal,
  title={Universal Kernels.},
  author={Micchelli, Charles A and Xu, Yuesheng and Zhang, Haizhang},
  journal={Journal of Machine Learning Research},
  volume={7},
  number={12},
  year={2006}
}

@inproceedings{hernandez2016predictive,
  title={Predictive entropy search for multi-objective bayesian optimization},
  author={Hern{\'a}ndez-Lobato, Daniel and Hernandez-Lobato, Jose and Shah, Amar and Adams, Ryan},
  booktitle={International conference on machine learning},
  pages={1492--1501},
  year={2016},
  organization={PMLR}
}

@article{belakaria2019max,
  title={Max-value entropy search for multi-objective Bayesian optimization},
  author={Belakaria, Syrine and Deshwal, Aryan and Doppa, Janardhan Rao},
  journal={Advances in neural information processing systems},
  volume={32},
  year={2019}
}

@article{volpp2019meta,
  title={Meta-learning acquisition functions for transfer learning in bayesian optimization},
  author={Volpp, Michael and Fr{\"o}hlich, Lukas P and Fischer, Kirsten and Doerr, Andreas and Falkner, Stefan and Hutter, Frank and Daniel, Christian},
  journal={arXiv preprint arXiv:1904.02642},
  year={2019}
}

@article{hsieh2021reinforced,
  title={Reinforced few-shot acquisition function learning for bayesian optimization},
  author={Hsieh, Bing-Jing and Hsieh, Ping-Chun and Liu, Xi},
  journal={Advances in Neural Information Processing Systems},
  volume={34},
  pages={7718--7731},
  year={2021}
}

@article{maraval2024end,
  title={End-to-end meta-Bayesian optimisation with transformer neural processes},
  author={Maraval, Alexandre and Zimmer, Matthieu and Grosnit, Antoine and Bou Ammar, Haitham},
  journal={Advances in Neural Information Processing Systems},
  volume={36},
  year={2024}
}

@article{zhang2025context,
  title={In-Context Multi-Objective Optimization},
  author={Zhang, Xinyu and Hassan, Conor and Martinelli, Julien and Huang, Daolang and Kaski, Samuel},
  journal={arXiv preprint arXiv:2512.11114},
  year={2025}
}

@book{evans2022partial,
  title={Partial differential equations},
  author={Evans, Lawrence C},
  volume={19},
  year={2022},
  publisher={American mathematical society}
}

@book{gilbarg1998elliptic,
  title={Elliptic partial differential equations of second order},
  author={Gilbarg, David and Trudinger, Neil S and Gilbarg, David and Trudinger, NS},
  volume={2},
  number={1},
  year={1998},
  publisher={Springer}
}

@article{hardy1916weierstrass,
  title={Weierstrass’s non-differentiable function},
  author={Hardy, Godefroy Harold},
  journal={Trans. Amer. Math. Soc},
  volume={17},
  number={3},
  pages={301--325},
  year={1916}
}

@incollection{takagi1990simple,
  title={A simple example of the continuous function without derivative},
  author={Takagi, Teiji},
  booktitle={Collected Papers},
  pages={5--6},
  year={1990},
  publisher={Springer}
}

@article{da2026sample,
  title={Sample path regularity of Gaussian processes from the covariance kernel},
  author={Da Costa, Natha{\"e}l and Pf{\"o}rtner, Marvin and Da Costa, Lancelot and Hennig, Philipp},
  journal={Analysis and Applications},
  pages={1--29},
  year={2026},
  publisher={World Scientific}
}


\clearpage
\appendix

\section{Model and Algorithm Details}
\label{apd:model_algorithm_details}

\subsection{Details of Foundation Model and Pre-training}
\label{apd:Details of Foundation Model and Pre-training}
\begin{table}[ht]
\centering
\caption{Detailed information for the foundation model and pre-training process, including the hyperparameters and corresponding values of the model architecture, data generation and model training.}\label{apd:tab:hyper of training}
\renewcommand{\arraystretch}{1.1}
\scalebox{0.8}{
\begin{tabular}{lll}
\toprule[1pt]
Type & Name  & Value  \\ \hline
\multirow{5}{*}{Model Architecture}
& Embedding Size & 512  \\
& Hidden Size & 1024 \\
& Attention Heads & 4 \\
& Transformer Layers & 12 \\
& Output Logits Size & 1000 \\
\hline
\multirow{8}{*}{Data Generation}
& GP Mean Prior & 0.0  \\
& GP Kernel & RBF  \\
& Output Scale & 1.0 \\
& Output Noise & $\mathcal N(0,10^{-4})$ \\ 
& Length Scale Prior & $\Gamma(\alpha=3.0,\beta=6.0)$ \\
& Maximum Feature Dimension & 30 \\
& Maximum Objective Dimension & 6 \\
& Maximum Sample Length & 256 \\
\hline
\multirow{5}{*}{Model Training}
& Batch Size & 128  \\
& Steps per Epoch & 2048  \\
& Linear-warmup Epochs & 5 \\
& Total Epochs & 500 \\
& Learning Rate & $5\times10^{-5}$ \\ 
\bottomrule[1pt]
\end{tabular}}
\end{table}

Our foundational model mainly employs the recently proposed Prior-Data Fitted Networks (PFNs) \cite{muller2021transformers} as the basic architecture, which is based on a Transformer encoder \cite{vaswani2017attention}. We utilize linear layers to encode trajectory and query points, along with sampled preferences, as contextual inputs for the foundation model. Following the methods outlined in \cite{hollmann2022tabpfn} and \cite{muller2023pfns4bo}, we set a maximum dimension $K$ for encoding the inputs, which are applicable to both feature and objective dimensions. We zero-pad the sampled dimension $k$ and linearly scale the dimension by a factor of $K/k$ to ensure that the magnitudes of the input encodings are consistent across different dimensions. Since different query points should not depend on one another, an attention mask is implemented to allow each trajectory point can only attend to each other, while the query points can only attend to the trajectory points. To maintain the permutation invariance of the input dataset, the positional encodings are removed and the input points are fed to the model as the sum of their encoded embeddings. During training, we maintain a maximum sample length of $N$ with $n$ trajectory points and $N-n$ query points, as used in \cite{muller2021transformers}. We sample each $n$ from $\{1,...,N-1\}$ with the weight of $1/(N-n)$ to simulate varying trajectory sizes during the optimization process, while ensuring that each trajectory size has approximately the same number of query points for training. To model the continuous distribution for prediction, the regression head of PFNs discretizes the continuous distribution into a piecewise constant distribution, referred to as the Riemann distribution \cite{muller2021transformers}. This helps transform the model training from a regression task to a classification problem. The boundaries of each discrete interval are estimated by sampling a large synthetic dataset from the same synthetic prior, ensuring that each interval maintains equal probability in the prior data. Additionally, the last bar on each side is replaced with a suitably scaled half-normal distribution to enhance training stability. To account for the distributional shifts in aggregated posteriors across varying objective dimensions, we propose an objective-aware regression head. This architecture dynamically routes latent representations to specialized regressor conditioned on the specific objective dimension of the problem. By leveraging this dimension-specific mapping, the model can more accurately capture the differences of the aggregated posteriors with varying objective dimensions, thereby significantly bolstering predictive precision when generalizing to unseen optimization landscapes across diverse objective dimensions.

During the pre-training phase, we extensively sample from a data generation mechanism to create a large and diverse set of synthetic data. Initially, we sample the number of features $d$, the number of objectives $m$, and the trajectory length $n$ from 1 up to their predetermined maximum values. We then sample $m$ independent latent functions from a Gaussian Process (GP) prior with zero mean and the radial basis function (RBF) kernel:
\begin{equation}
f \sim \mathcal{G} \mathcal{P}(0, k_\textnormal{RBF}(\cdot,\cdot)), \quad k_{\textnormal{RBF}}(\boldsymbol{x}, \boldsymbol{x}^{\prime})=\sigma^2 \exp (-\frac{\|\boldsymbol{x}-\boldsymbol{x}^{\prime}\|^2}{2 l^2}),
\end{equation}
where $\sigma$ and $l$ represent the output scale and length scale kernel parameters, respectively. We also sample the length scale parameter for each feature dimension during the latent function sampling, along with a random observation noise to enhance function diversity. In the third step, we randomly sample input features from the $d$-dimensional unit cube, and then evaluate the sampled latent functions to generate the corresponding observations, resulting in $n$ trajectory samples and $N-n$ query samples. After that, we randomly sample the artificial preferences from the simplex $\Delta=\{\boldsymbol{\lambda}\in\mathbb{R}_{+}^{m}\mid\|\boldsymbol{\lambda}\|_{1}=1\}$ to derive the aggregation targets $g$ given query inputs and preferences. The aggregation targets are then masked to formulate a supervised prediction problem, enabling the model to learn to predict the preference-wise aggregated posterior distribution $q_{\boldsymbol{\theta}}(g|\boldsymbol{x},D_n;\boldsymbol{\lambda})$. We continuously synthesize data through extensive sampling and optimize the model parameters to minimize the cross-entropy loss function defined in Equation (\ref{eq:loss function}), which is equivalent to the expected KL divergence between the true posterior distribution and its approximation, up to an additive constant \cite{muller2021transformers}. The detailed information and hyperparameter settings regarding the foundational model and the pre-training process can be found in Table \ref{apd:tab:hyper of training}.

\subsection{Visualization of Aggregated Posteriors}
\label{apd:Visualization of Aggregated Posteriors}
\begin{figure*}[!ht]
\centering
\includegraphics[scale=0.27]{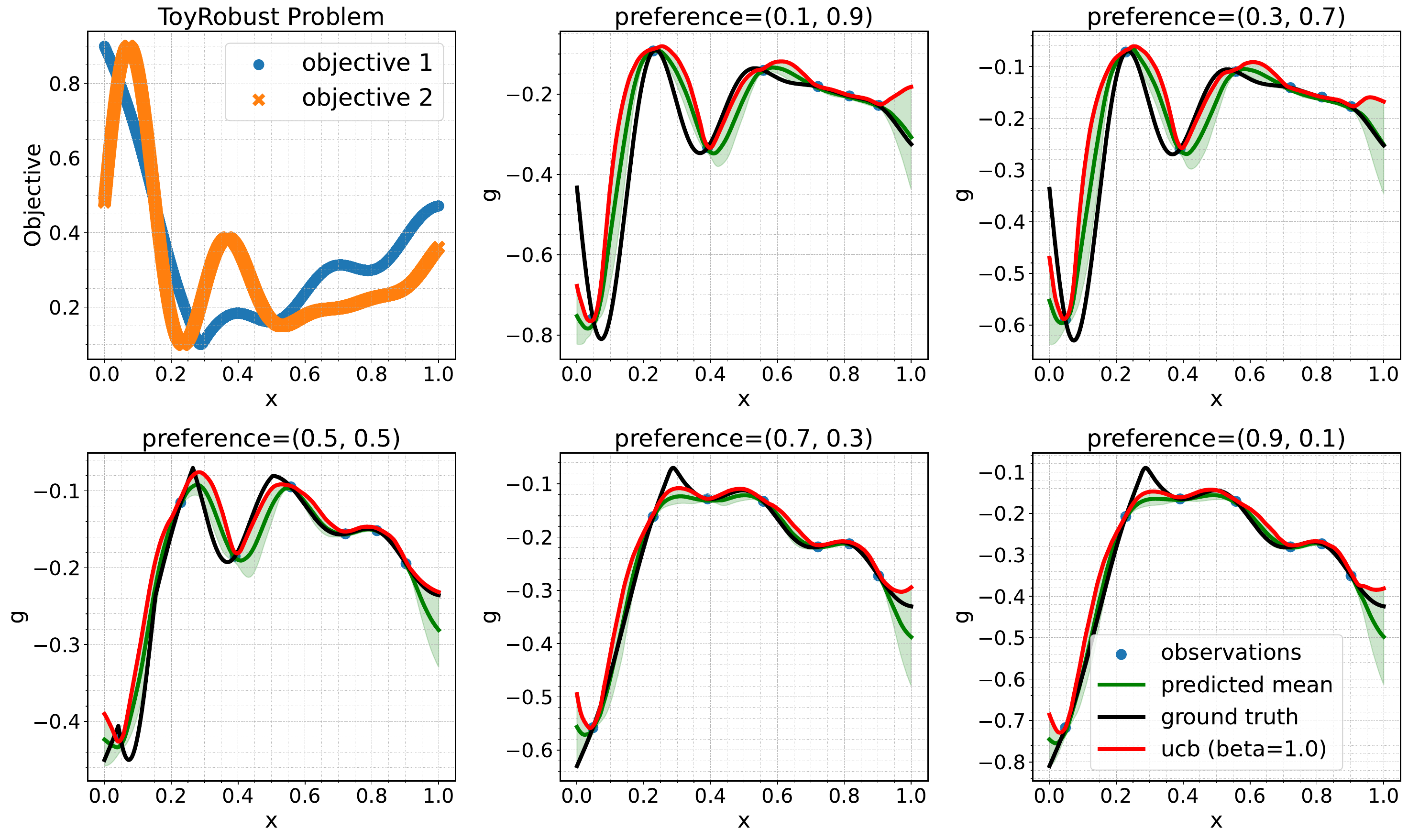}
\caption{Details of the normalized objective functions for the ToyRobust problem, the aggregated posteriors (green solid lines with shaded regions) conditioned on 7 observations and 5 preferences, and the corresponding UCB acquisition functions (red solid lines).}
\label{fig:aggregated posteriors}
\end{figure*}
To facilitate visualization, we employ a one-dimensional ToyRobust problem \cite{daulton2022robust} as a representative case to demonstrate the foundation model's ability to predict aggregated posteriors with unseen objective functions, as illustrated in Figure \ref{fig:aggregated posteriors}. The results indicate that the model provides reliable posterior estimates directly from the evaluated trajectory of the unknown problem, without requiring further training or parameter updates. This underscores the model's generalization capabilities beyond its training distribution. Specifically, the predicted mean accurately captures the global trends of individual sub-problems and also maintains similar local characteristics defined by adjacent preferences. Furthermore, the model yields well-calibrated uncertainty estimates, enabling acquisition functions, such as the Upper Confidence Bound (UCB) shown in the figure, to effectively balance exploration and exploitation throughout the optimization process.

\subsection{Derivation of UHVI Acquisition Function}
\label{apd:UHVI}
\paragraph{Definitions and preliminaries.} The derivation mainly depends on two types of definitions on Tchebycheff scalarizations and the theorem on hypervolume scalarization. We consider the following multi-objective optimization problem (MOP):
\begin{equation}
\min _{\boldsymbol{x} \in \mathcal{X}} \boldsymbol{f}(\boldsymbol{x})=(f_1(\boldsymbol{x}), f_2(\boldsymbol{x}), \cdots, f_m(\boldsymbol{x})).
\end{equation}
We denote $D_n=\{\boldsymbol{x}_{i},\boldsymbol{y}_{i}\}_{i=1}^{n}$ be a set of $n$ evaluated samples, where each $\boldsymbol{x}=(x_{1}, x_{2}, \ldots, x_{d})^T$ is the feature vector in the decision space $\mathcal{X} \subset \mathbb{R}^d$, and each $\boldsymbol{y}=(y_{1}, y_{2}, \ldots, y_{m})^T$ is the observation vector of the true objective function $\boldsymbol{f}(\boldsymbol{x})$. We denote $\boldsymbol{z}^*=(z_1^*, \cdots, z_m^*)^T$ as the ideal point (lower bound) in the objective space. We have the Tchebycheff scalarization function:
\begin{equation}
s_{\boldsymbol{\lambda}}^{\text{tch}}(\boldsymbol{x})=\max _{1 \leq j \leq m}\{\lambda_j(y_j(\boldsymbol{x})-z_j^*)\},
\label{eq:s_tch_lmd}
\end{equation}
where $\boldsymbol{\lambda}=(\lambda_{1}, \lambda_{2}, \ldots, \lambda_{m})^T$ represents a preference sampled from the simplex $\Delta=\{\boldsymbol{\lambda}\in\mathbb{R}_{+}^{m}\mid\|\boldsymbol{\lambda}\|_{1}=1\}$, we can identify each Pareto optimal solution by minimizing $s_{\boldsymbol{\lambda}}^{\text{tch}}(\boldsymbol{x})$ with a specific but unknown preference \cite{choo1983proper}.

The Tchebycheff scalarization function can be transformed in another formulation with respect to a set of reference
vectors $\boldsymbol{w}=(w_{1}, w_{2}, \ldots, w_{m})^T$ sampled from $S=\{\boldsymbol{w}\in\mathbb{R}_{+}^{m}\mid\|\boldsymbol{w}\|_{2}=1\}$:
\begin{equation}
s_{\boldsymbol{w}}^{\text{tch}}(\boldsymbol{x})=\max _{1 \leq j \leq m}\{(y_j(\boldsymbol{x})-z_j^*)/w_j\}.
\label{eq:s_tch_w}
\end{equation}
Leveraging Equation (\ref{eq:s_tch_w}), we have the following theorem, which will help us approximate the hypervolume indicator and further derive UHVI acquisition function.

\begin{theorem}[Hypervolume Scalarization]
\label{tr:HV scalar}
Let $D_n=\{\boldsymbol{x}_{i},\boldsymbol{y}_{i}\}_{i=1}^{n}$ be a set of $n$ evaluated samples, $\boldsymbol{z}^*=(z_1^*, \cdots, z_m^*)^T$ is the ideal point, $\boldsymbol{r}=(r_1, \cdots, r_m)^T$ is a pre-defined reference point. The hypervolume of the evaluated samples can be expressed as an expectation of Tchebycheff scalarization function over a set of uniformly sampled reference vectors $S=\{\boldsymbol{w}\in\mathbb{R}_{+}^{m}\mid\|\boldsymbol{w}\|_{2}=1\}$ \cite{deng2019approximating,zhang2020random,zhang2023hypervolume}:
\begin{equation}
\label{eq:HV scalar}
\textnormal{HV}=\prod_{j=1}^m(r_j-z_j^*)-c_m\mathbb{E}_{\boldsymbol{w}\sim S}\{[\min _{1 \leq i \leq n} s_{\boldsymbol{w}}^{\textnormal{tch}}(\boldsymbol{x}_i)]^m\},
\end{equation}
where $c_m=\frac{\pi^{m / 2}}{2^m \Gamma(m / 2+1)}$ is a constant that depends only on $m$.
\end{theorem}
Utilizing the hypervolume scalarization and Equation (\ref{eq:s_tch_lmd}), we can naturally derive the utility of UHVI acquisition function:

\begin{lemma}[Uncertainty-aware Hypervolume Improvement]
Given the distribution of the aggregation target $g_{\boldsymbol{\lambda}}(\cdot)=-s_{\boldsymbol{\lambda}}^{\textnormal{tch}}(\cdot)$ conditioned on evaluated samples $D_n=\{\boldsymbol{x}_{i},\boldsymbol{y}_{i}\}_{i=1}^{n}$, the utility of hypervolume improvement in the normalized objective space at any query point $\boldsymbol{x}$ can be estimated by the following expectation:
\begin{equation}
\alpha_{\textnormal{UHVI}}(\boldsymbol{x}) = c_m\mathbb{E}_{\boldsymbol{\lambda}\sim \Delta}[(c_{\boldsymbol{\lambda}})^m\max\{0, [\min _{1 \leq i \leq n}\{-g_{\boldsymbol{\lambda}}(\boldsymbol{x}_i)\}]^m
-[-\textnormal{UCB}(g_{\boldsymbol{\lambda}}(\boldsymbol{x}))]^m\}],
\end{equation}

where $c_m=\frac{\pi^{m / 2}}{2^m \Gamma(m / 2+1)}$ is a constant that depends only on $m$, $c_{\boldsymbol{\lambda}}=\sqrt{\sum_{j=1}^{m} \frac{1}{\lambda_j^2}}$ is a transformation constant that depends only on $\boldsymbol{\lambda}$, $\textnormal{UCB}(\cdot)$ denotes the upper confidence bound (UCB) metric.
\end{lemma}

\begin{proof}
From Equation (\ref{eq:HV scalar}), given an updated evaluated samples $D_{n+1}=\{\boldsymbol{x}_{i},\boldsymbol{y}_{i}\}_{i=1}^{n+1}$ with newly added sample $\{\boldsymbol{x}_{n+1},\boldsymbol{y}_{n+1}\}$, the hypervolume can be represented as:
\begin{equation}
\textnormal{HV}'=\prod_{j=1}^m(r_j-z_j^*)-c_m\mathbb{E}_{\boldsymbol{w}\sim S}\{[\min _{1 \leq i \leq n+1} s_{\boldsymbol{w}}^{\textnormal{tch}}(\boldsymbol{x}_i)]^m\},
\end{equation}
which induces the hypervolume improvement:
\begin{equation}
\begin{aligned}
\textnormal{HVI}(\boldsymbol{x}_{n+1}) &= \textnormal{HV}'-\textnormal{HV}\\
&= c_m\mathbb{E}_{\boldsymbol{w}\sim S}\{[\min _{1 \leq i \leq n} s_{\boldsymbol{w}}^{\textnormal{tch}}(\boldsymbol{x}_i)]^m-[\min _{1 \leq i \leq n+1} s_{\boldsymbol{w}}^{\textnormal{tch}}(\boldsymbol{x}_i)]^m\}\\
&= c_m\mathbb{E}_{\boldsymbol{w}\sim S}\{[\min _{1 \leq i \leq n} s_{\boldsymbol{w}}^{\textnormal{tch}}(\boldsymbol{x}_i)]^m-[\min\{\min _{1 \leq i \leq n} s_{\boldsymbol{w}}^{\textnormal{tch}}(\boldsymbol{x}_i),s_{\boldsymbol{w}}^{\textnormal{tch}}(\boldsymbol{x}_{n+1})\}]^m\}\\
&= c_m\mathbb{E}_{\boldsymbol{w}\sim S}[\max\{0,[\min _{1 \leq i \leq n} s_{\boldsymbol{w}}^{\textnormal{tch}}(\boldsymbol{x}_i)]^m-[s_{\boldsymbol{w}}^{\textnormal{tch}}(\boldsymbol{x}_{n+1})]^m\}].
\end{aligned}
\end{equation}
Given the transformation constant $c_{\boldsymbol{\lambda}}=\sqrt{\sum_{j=1}^{m} \frac{1}{\lambda_j^2}}$, each reference vector $\boldsymbol{w}$ can be uniquely represented by the preference $\boldsymbol{\lambda}$ by using $w_j=1/({\lambda_j c_{\boldsymbol{\lambda}})}$, the hypervolume improvement can be represented as:
\begin{equation}
\textnormal{HVI}(\boldsymbol{x}_{n+1}) = c_m\mathbb{E}_{\boldsymbol{\lambda}\sim \Delta}[(c_{\boldsymbol{\lambda}})^m\max\{0,[\min _{1 \leq i \leq n} s_{\boldsymbol{\lambda}}^{\textnormal{tch}}(\boldsymbol{x}_i)]^m-[s_{\boldsymbol{\lambda}}^{\textnormal{tch}}(\boldsymbol{x}_{n+1})]^m\}].
\end{equation}
Our model predict the distribution of the aggregation target $g_{\boldsymbol{\lambda}}(\boldsymbol{x}_{i})=-s_{\boldsymbol{\lambda}}^{\textnormal{tch}}(\boldsymbol{x}_{i})$ given each $\boldsymbol{x}_{i}$. To ensure that objectives of varying magnitudes share the same scale, we normalize the objectives to the range of [0,1] during the optimization process. Consequently, the hypervolume improvement in the normalized objective space can be reformulated as follows:
\begin{equation}
\textnormal{HVI}(\boldsymbol{x}_{n+1}) = c_m\mathbb{E}_{\boldsymbol{\lambda}\sim \Delta}[(c_{\boldsymbol{\lambda}})^m\max\{0,[\min _{1 \leq i \leq n} \{-g_{\boldsymbol{\lambda}}(\boldsymbol{x}_i)\}]^m-[-g_{\boldsymbol{\lambda}}(\boldsymbol{x}_{n+1})]^m\}].
\end{equation}
Here the aggregation target $g_{\boldsymbol{\lambda}}(\cdot)$ is constrained to be non-positive after the objective normalization (see Figure \ref{fig:aggregated posteriors} for an example). Since we wish to maximize the aggregation target at the query point $\boldsymbol{x}_{n+1}$ for maximizing hypervolume improvement, we replace the prediction $g_{\boldsymbol{\lambda}}(\boldsymbol{x}_{n+1})$ with the $\textnormal{UCB}(\cdot)$ metric as an upper bound estimation, which introduces the calibrated uncertainty provided by our model for searching the candidates with a good balance of exploration and exploitation:
\begin{equation}
\hat{\textnormal{HVI}}(\boldsymbol{x}_{n+1}) = c_m\mathbb{E}_{\boldsymbol{\lambda}\sim \Delta}[(c_{\boldsymbol{\lambda}})^m\max\{0,[\min _{1 \leq i \leq n} \{-g_{\boldsymbol{\lambda}}(\boldsymbol{x}_i)\}]^m-[-\textnormal{UCB}(g_{\boldsymbol{\lambda}}(\boldsymbol{x}_{n+1}))]^m\}].
\end{equation}
We can then derive the utility of UHVI acquisition function as $\boldsymbol{x}_{n+1}$ is arbitrary:
\begin{equation}
\alpha_{\textnormal{UHVI}}(\boldsymbol{x}) = c_m\mathbb{E}_{\boldsymbol{\lambda}\sim \Delta}[(c_{\boldsymbol{\lambda}})^m\max\{0, [\min _{1 \leq i \leq n}\{-g_{\boldsymbol{\lambda}}(\boldsymbol{x}_i)\}]^m
-[-\textnormal{UCB}(g_{\boldsymbol{\lambda}}(\boldsymbol{x}))]^m\}],
\end{equation}
which leads to our claim in the lemma.
\qedhere
\end{proof}
Intuitively, maximizing $\alpha_{\textnormal{UHVI}}$ utility is equivalent to maximizing the predicted upper bounds of aggregation targets across all preference simultaneously. We find that our utility exhibits similar expression to that presented in \cite{song2024vizier}. However, the key difference is that the formulation of \cite{song2024vizier} is based on the Gaussian process (GP) posterior for each objective, whereas our acquisition function is directly derived from the infinite preference-wise aggregated posteriors, which is intractable for traditional GP training and updates.

\subsection{Derivation of UR2I Acquisition Function}
\begin{theorem}[R2 Indicator]
\label{tr:R2 scalar}
Let $D_n=\{\boldsymbol{x}_{i},\boldsymbol{y}_{i}\}_{i=1}^{n}$ be a set of $n$ evaluated samples, the R2 indicator of the evaluated samples is defined as an expectation of Tchebycheff scalarization function over a set of preferences sampled from the simplex $\Delta=\{\boldsymbol{\lambda}\in\mathbb{R}_{+}^{m}\mid\|\boldsymbol{\lambda}\|_{1}=1\}$ \cite{hansen1994evaluating,brockhoff2012properties,zhao2024many}:
\begin{equation}
\label{eq:R2 scalar}
\textnormal{R}_{2}=\mathbb{E}_{\boldsymbol{\lambda}\sim \Delta}[\min _{1 \leq i \leq n} s_{\boldsymbol{\lambda}}^{\textnormal{tch}}(\boldsymbol{x}_i)].
\end{equation}
\end{theorem}
Analogous to the derivation of the UHVI acquisition function presented above, the utility for the UR2I acquisition function can be naturally derived as follows. For the sake of brevity, the formal proof is omitted here.

\begin{lemma}[Uncertainty-aware R2 Improvement]
Given the distribution of the aggregation target $g_{\boldsymbol{\lambda}}(\cdot)=-s_{\boldsymbol{\lambda}}^{\textnormal{tch}}(\cdot)$ conditioned on evaluated samples $D_n=\{\boldsymbol{x}_{i},\boldsymbol{y}_{i}\}_{i=1}^{n}$, the utility of R2 improvement in the normalized objective space at any query point $\boldsymbol{x}$ can be estimated by the following expectation:
\begin{equation}
\alpha_{\textnormal{UR2I}}(\boldsymbol{x}) =\mathbb{E}_{\boldsymbol{\lambda}\sim \Delta}[\max\{0, \min _{1 \leq i \leq n}\{-g_{\boldsymbol{\lambda}}(\boldsymbol{x}_i)\}
-[-\textnormal{UCB}(g_{\boldsymbol{\lambda}}(\boldsymbol{x}))]\}].
\end{equation}
\end{lemma}

\section{Experimental Setup Details}
\label{apd:experimental setup details}

\subsection{Benchmark Problems}
\label{apd:Benchmark Problems}
\begin{table}[ht]
\centering
\caption{Detailed information of test benchmarks.}\label{apd:tab:benchmarks}
\renewcommand{\arraystretch}{1.1}
\scalebox{0.8}{
\begin{tabular}{llcc}
\toprule[1pt]
Type & Name  & Feature Dimension ($d$)  & Objective Dimension ($m$)  \\ 
\hline

\multirow{6}{*}{Synthetic Function} 
& BC (Branin-Currin) &2 & 2 \\
& AR (Ackley-Rosenbrock) &2 & 2 \\
& ABC (Ackley-Branin-Currin) &2 & 3 \\
& ABD (Ackley-Branin-Dixon) &2 & 3 \\
& ACD (Ackley-Currin-Dixon) &2 & 3 \\
& BCD (Branin-Currin-Dixon) &2 & 3 \\
\hline
\multirow{10}{*}{Engineering Design} 
& Four Bar Truss Design &4 & 2 \\
& Pressure Vessel Design &4 & 2 \\
& Hatch Cover Design &2 & 2 \\
& Disc Brake Design &4 & 3 \\
& Speed Reducer Design &7 & 3 \\
& Gear Train Design &4 & 3 \\
& Rocket Injector Design &4 & 3 \\
& Car Side Impact Design &7 & 4 \\
& Conceptual Marine Design &6 & 4 \\
& Water Resource Planning &3 & 6 \\
\hline
\multirow{4}{*}{Hyperparameter Optimization} 
& Lego &5 & 2 \\
& Materials &5 & 2 \\
& Mic &5 & 3 \\
& ship &5 & 3 \\
\bottomrule[1pt]
\end{tabular}}
\end{table}

To evaluate the performance and generalization of our method, we employ a diverse suite of synthetic and real-world benchmarks, including 6 synthetic functions, 10 real-world engineering design problems and 4 hyperparameter optimization tasks. Specifically, we firstly utilize a combination of synthetic black-box functions characterized by a wide spectrum of landscapes, including unimodal versus multimodal structures, smooth versus non-smooth surfaces, and challenges involving narrow valleys and complex variable coupling, we utilize the implementation in BOFormer \cite{hung2025boformer}. Secondly, we incorporate the RE benchmark suite \cite{tanabe2020easy}, which comprises real-world multi-objective engineering design problems from various fields such as pressure vessel, speed reducer, and rocket injector design, covering multi-objective ($m$ = 2, 3) to many-objective ($m$ = 4, 6) scenarios. Detailed description and characterization of the objective functions for these problems can be found in \cite{tanabe2020easy}. Finally, we validate our approach on a hyperparameter optimization (HPO) problem for 3D Gaussian Splatting (3DGS). This problem aims to identify the optimal configurations of the the recent state-of-the-art method 3DGS \cite{kerbl20233d} in 3D object reconstruction task \cite{mildenhall2021nerf}. The optimization objectives encompass rendering quality (measured by PSNR), model size, and the total number of Gaussian primitives. To comprehensively assess performance, we conduct evaluations across four diverse reconstruction scenes: Lego, Materials, Mic, and Ship from \citep{mildenhall2021nerf}. The detailed information of the benchmarks can be referred to Table \ref{apd:tab:benchmarks}. 

\subsection{Algorithm Hyperparameters}
\label{apd:Algorithm Hyperparameters}
Since qNEHVI, qParEGO, JES, and our proposed method are all implemented based on the BoTorch framework \cite{balandat2020botorch}, we maintain consistent optimization settings for their respective acquisition functions. Specifically, we employ the multi-start L-BFGS-B optimizer with 20 restarts seeded from 1024 pseudo-random samples through BoTorch’s initialization heuristic, consistent with the protocol in \cite{ament2023unexpected}. To strike a balance between optimization accuracy and computational efficiency, we utilize a stochastic approximation for the UHVI and UR2I utilities by randomly sampling 100 preference vectors during each optimization step. We conduct ablation study on number of preferences in Appendix \ref{apd:UHVI pref ablation}. For all other baselines, we adhere to the default parameter configurations provided in their original code implementations unless otherwise specified.

\subsection{Performance Metrics}
\label{apd:Performance Metrics}
To rigorously evaluate the quality of the Pareto fronts obtained by different algorithms, we employ the widely-used hypervolume (HV) indicator \cite{zitzler2002multiobjective} as the performance metric. For both synthetic black-box functions and real-world hyperparameter optimization tasks, we adopt the evaluation protocol from BOFormer \cite{hung2025boformer}, in which the true objective values are normalized using the pre-defined bounds. Consequently, we set the reference point to $\boldsymbol{r}=[1.1,...,1.1]^T\in\mathbb{R}^{m}$ for HV computation in the normalized objective space. For real-world engineering design problems, we utilize the approximate Pareto fronts provided by \cite{tanabe2020easy} as the ground truth, which are generated with a large number of evaluations and have also been widely used in other related studies. To ensure fairness, we firstly generate the ideal point $\boldsymbol{y}_{\text {ideal}}$ and nadir point $\boldsymbol{y}_{\text {nadir}}$ from the approximate Pareto front of the tested problem:
\begin{equation}
\begin{aligned}
&\boldsymbol{y}_{\text {ideal}}=(\min y_{1}(\boldsymbol{x}_{1}), \ldots, \min y_{m}(\boldsymbol{x}_{m}))^T, \\
&\boldsymbol{y}_{\text {nadir}}=(\max y_{1}(\boldsymbol{x}_{1}), \ldots, \max y_{m}(\boldsymbol{x}_{m}))^T, \\
&\forall \boldsymbol{x}_{1}, \boldsymbol{x}_{2}, \ldots, \boldsymbol{x}_{m} \in \mathcal{P}_s,
\end{aligned}
\end{equation}
where $\boldsymbol{x}_{1}, \boldsymbol{x}_{2}, \ldots, \boldsymbol{x}_{m}$ are solutions in approximate Pareto set $\mathcal{P}_s$ from the ground truth. After that, we normalize the resulting Pareto fronts obtained from the algorithms using the bounds of ideal point and nadir point. Finally, we also set the reference point as $\boldsymbol{r}=[1.1,...,1.1]^T\in\mathbb{R}^{m}$ to calculate the hypervolume indicator of the normalized Pareto front obtained by each algorithm.

\section{Complete Performance Results}

\label{apd:complete performance results}

Tables~\ref{tab:bbob exp},~\ref{tab:re exp}, and~\ref{tab:hpo exp} present the complete sequential optimization performance results on synthetic black-box functions, engineering design problems, and hyperparameter optimization tasks, respectively.

\begin{table}[htbp]
\centering
\caption{Means (stds) of hypervolume metrics ($\uparrow$) on synthetic black-box functions over 10 independent runs. The best, second-best, and third-best mean results for each problem are highlighted with dark gray, gray, and light gray backgrounds, respectively.}\label{tab:bbob exp}
\renewcommand{\arraystretch}{1.5}
\scalebox{0.56}{
\begin{tabular}{ccccccc}
\toprule[1pt]
Algorithm & BC & AR & ABC & ABD & ACD & BCD \\
\hline
MOEA/D-EGO & 7.015e-01(7.5e-03) & \cellcolor{gray!30}\textbf{1.203e+00(2.4e-03)} & 4.826e-01(8.1e-03) & 6.934e-01(1.1e-02) & 9.320e-01(1.4e-02) & 5.735e-01(1.9e-02) \\
TSEMO & 8.056e-01(9.1e-04) & 1.172e+00(1.4e-02) & 4.853e-01(4.5e-03) & 7.114e-01(9.4e-03) & 9.117e-01(2.6e-02) & 6.938e-01(1.1e-02) \\
USEMO-EI & 7.750e-01(1.2e-02) & 1.118e+00(2.3e-02) & 4.648e-01(8.2e-03) & 7.298e-01(1.1e-02) & 8.991e-01(2.5e-02) & 6.766e-01(9.5e-03) \\
DGEMO & \cellcolor{gray!50}\textbf{8.087e-01(1.3e-03)} & 1.158e+00(6.8e-02) & 4.477e-01(3.8e-02) & 5.918e-01(6.8e-02) & 7.266e-01(9.1e-02) & 7.054e-01(3.9e-03) \\
qNEHVI & 7.962e-01(1.7e-03) & 1.159e+00(1.8e-02) & \cellcolor{gray!10}\textbf{4.929e-01(6.5e-03)} & 7.181e-01(1.0e-02) & 9.167e-01(1.3e-02) & 6.858e-01(1.5e-03) \\
qParEGO & 7.843e-01(4.2e-03) & 1.194e+00(4.4e-03) & 4.782e-01(7.5e-03) & 7.032e-01(9.1e-03) & 9.285e-01(1.3e-02) & 7.041e-01(4.5e-03) \\
JES & 7.765e-01(3.1e-03) & 1.086e+00(3.4e-02) & 4.559e-01(1.1e-02) & 6.598e-01(1.3e-02) & 8.802e-01(2.8e-02) & 6.947e-01(3.5e-03) \\
PSL & 6.708e-01(1.8e-02) & 6.375e-01(1.6e-01) & 2.785e-01(4.6e-02) & 4.548e-01(8.9e-02) & 5.027e-01(1.2e-01) & 6.445e-01(1.0e-02) \\
BOFormer & 7.187e-01(1.4e-02) & 1.092e+00(3.4e-02) & 4.082e-01(1.7e-02) & 6.383e-01(2.5e-02) & 7.988e-01(3.8e-02) & 6.715e-01(1.5e-02) \\
TAMO & 7.411e-01(8.7e-03) & \cellcolor{gray!50}\textbf{1.210e+00}(7.5e-06) & 4.904e-01(4.3e-03) & 7.270e-01(4.3e-03) & \cellcolor{gray!50}\textbf{9.723e-01}(3.1e-03) & 6.853e-01(4.1e-03) \\
\hline
FoMEMO-EI & \cellcolor{gray!10}\textbf{8.066e-01(6.2e-04)} & 1.192e+00(1.0e-02) & \cellcolor{gray!50}\textbf{4.993e-01(1.2e-02)} & \cellcolor{gray!10}\textbf{7.308e-01(7.9e-03)} & 9.291e-01(1.5e-02) & \cellcolor{gray!30}\textbf{7.346e-01(2.3e-03)} \\
FoMEMO-UCB & \cellcolor{gray!30}\textbf{8.075e-01(5.8e-04)} & 1.193e+00(8.4e-03) & 4.878e-01(1.4e-02) & \cellcolor{gray!30}\textbf{7.369e-01(8.5e-03)} & \cellcolor{gray!10}\textbf{9.407e-01(1.5e-02)} & 7.294e-01(6.2e-03) \\
FoMEMO-UHVI & 8.052e-01(4.9e-04) & \cellcolor{gray!10}\textbf{1.195e+00(6.3e-03)} & 4.908e-01(7.4e-03) & \cellcolor{gray!50}\textbf{7.430e-01(8.4e-03)} & \cellcolor{gray!30}\textbf{9.519e-01(1.2e-02)} & \cellcolor{gray!50}\textbf{7.483e-01(7.6e-04)} \\
FoMEMO-UR2I & 8.016e-01(1.7e-03) & 1.164e+00(2.2e-02) & \cellcolor{gray!30}\textbf{4.973e-01(6.0e-03)} & 7.126e-01(9.1e-03) & 9.256e-01(1.8e-02) & \cellcolor{gray!10}\textbf{7.303e-01(1.2e-03)} \\
\bottomrule[1pt]
\end{tabular}}
\end{table}

\begin{table}[htbp]
\centering
\caption{Means (stds) of hypervolume metrics ($\uparrow$) on engineering design problems over 10 independent runs. The best, second-best, and third-best mean results for each problem are highlighted with dark gray, gray, and light gray backgrounds, respectively.}\label{tab:re exp}
\renewcommand{\arraystretch}{1.5}
\scalebox{0.53}{
\begin{tabular}{cccccccc}
\toprule[1pt]
Algorithm & Four Bar Truss & Pressure Vessel & Hatch Cover & Disc Brake & Speed Reducer & Gear Train & Rocket Injector \\
\hline
MOEA/D-EGO & 7.462e-01(1.4e-02) & 1.129e+00(8.8e-03) & 1.119e+00(2.3e-02) & 1.280e+00(1.4e-02) & 1.261e+00(1.3e-02) & 5.623e-01(1.3e-01) & 7.539e-01(1.1e-02) \\
TSEMO & 8.506e-01(4.7e-04) & 1.099e+00(1.6e-02) & 1.022e+00(3.5e-02) & 1.285e+00(3.0e-02) & 1.297e+00(1.3e-03) & 8.092e-01(2.3e-02) & \cellcolor{gray!10}\textbf{8.618e-01(7.5e-04)} \\
USEMO-EI & 8.232e-01(4.3e-03) & \cellcolor{gray!50}\textbf{1.153e+00(1.6e-03)} & 1.145e+00(6.2e-03) & \cellcolor{gray!10}\textbf{1.305e+00(2.3e-03)} & 1.286e+00(3.7e-03) & 6.433e-01(4.0e-02) & 8.316e-01(3.7e-03) \\
DGEMO & 8.526e-01(9.8e-04) & 1.132e+00(6.3e-03) & 1.003e+00(7.9e-02) & 1.286e+00(2.9e-02) & \cellcolor{gray!50}\textbf{1.304e+00(1.6e-04)} & 8.859e-01(3.2e-02) & 8.605e-01(1.9e-03) \\
qNEHVI & 8.244e-01(3.8e-03) & 1.069e+00(1.6e-02) & 1.119e+00(1.3e-02) & 1.192e+00(3.1e-02) & 1.263e+00(7.9e-03) & 8.387e-01(2.6e-02) & 8.182e-01(5.0e-03) \\
qParEGO & 7.917e-01(1.7e-02) & 1.099e+00(1.5e-02) & 1.144e+00(6.7e-03) & 1.199e+00(7.4e-02) & 1.215e+00(1.4e-02) & 7.813e-01(3.1e-02) & 6.700e-01(2.8e-02) \\
JES & 7.918e-01(1.0e-02) & 8.795e-01(1.3e-01) & \cellcolor{gray!10}\textbf{1.150e+00(3.5e-03)} & 1.249e+00(2.5e-02) & 1.206e+00(1.0e-02) & 6.096e-01(4.5e-02) & 6.917e-01(2.4e-02) \\
PSL & 8.655e-01(2.8e-03) & 4.971e-02(1.0e-02) & 9.752e-01(9.6e-04) & 1.153e+00(3.7e-02) & 1.053e+00(2.4e-02) & 8.488e-01(2.2e-02) & 6.191e-01(2.4e-02) \\
BOFormer & 7.308e-01(1.4e-02) & 3.094e-01(1.7e-01) & 1.119e+00(1.5e-02) & 1.166e+00(2.3e-02) & 1.191e+00(1.0e-02) & 3.818e-01(5.7e-02) & 6.208e-01(2.2e-02) \\
TAMO & 8.069e-01(3.9e-03) & 6.643e-01(7.9e-02) & 1.150e+00(2.6e-03) & 1.253e+00(1.7e-03) & N/A & 6.460e-01(8.4e-05) & 7.452e-01(2.9e-03) \\
\hline
FoMEMO-EI & \cellcolor{gray!10}\textbf{8.667e-01(2.0e-03)} & \cellcolor{gray!30}\textbf{1.151e+00(1.8e-03)} & \cellcolor{gray!30}\textbf{1.162e+00(1.5e-03)} & \cellcolor{gray!30}\textbf{1.309e+00(2.0e-03)} & \cellcolor{gray!10}\textbf{1.298e+00(7.3e-04)} & \cellcolor{gray!30}\textbf{9.107e-01(8.1e-03)} & 8.506e-01(4.6e-03) \\
FoMEMO-UCB & \cellcolor{gray!30}\textbf{8.722e-01(1.2e-03)} & \cellcolor{gray!10}\textbf{1.150e+00(1.8e-03)} & \cellcolor{gray!50}\textbf{1.163e+00(1.9e-03)} & \cellcolor{gray!50}\textbf{1.312e+00(3.1e-04)} & \cellcolor{gray!30}\textbf{1.299e+00(6.2e-04)} & \cellcolor{gray!50}\textbf{9.270e-01(4.5e-03)} & \cellcolor{gray!30}\textbf{8.666e-01(2.4e-03)} \\
FoMEMO-UHVI & 8.609e-01(1.3e-03) & 1.022e+00(5.4e-02) & 1.127e+00(1.1e-02) & 1.273e+00(1.6e-02) & 1.263e+00(3.7e-03) & 8.573e-01(9.6e-03) & 8.245e-01(7.3e-03) \\
FoMEMO-UR2I & \cellcolor{gray!50}\textbf{8.741e-01(5.0e-04)} & 1.086e+00(2.3e-02) & 1.131e+00(4.8e-03) & 1.299e+00(7.2e-03) & 1.285e+00(3.6e-03) & \cellcolor{gray!10}\textbf{8.961e-01(1.1e-02)} & \cellcolor{gray!50}\textbf{8.693e-01(1.5e-03)} \\
\bottomrule[1pt]
\end{tabular}}
\end{table}

\begin{table}[htbp]
\centering
\caption{Means (stds) of hypervolume metrics ($\uparrow$) on hyperparameter optimization problems over 10 independent runs. The best, second-best, and third-best mean results for each problem are highlighted with dark gray, gray, and light gray backgrounds, respectively.}\label{tab:hpo exp}
\renewcommand{\arraystretch}{1.5}
\scalebox{0.56}{
\begin{tabular}{ccccc}
\toprule[1pt]
Algorithm & Lego & Materials & Mic & Ship \\
\hline
MOEA/D-EGO & \cellcolor{gray!50}\textbf{1.184e+00(1.1e-03)} & 1.166e+00(2.2e-03) & 1.257e+00(4.9e-03) & \cellcolor{gray!50}\textbf{1.321e+00(1.5e-03)} \\
TSEMO & 1.174e+00(3.3e-03) & 1.160e+00(5.1e-03) & 1.251e+00(4.8e-03) & 1.315e+00(2.1e-03) \\
USEMO-EI & \cellcolor{gray!10}\textbf{1.179e+00(1.6e-03)} & 1.167e+00(2.8e-03) & 1.255e+00(5.4e-03) & 1.315e+00(3.6e-03) \\
DGEMO & 1.171e+00(5.0e-03) & 1.172e+00(3.2e-03) & \cellcolor{gray!30}\textbf{1.273e+00(2.1e-02)} & 1.313e+00(3.6e-03) \\
qNEHVI & 1.147e+00(9.8e-03) & 1.144e+00(1.1e-02) & 1.198e+00(1.9e-02) & 1.293e+00(6.6e-03) \\
qParEGO & 1.137e+00(1.2e-02) & 1.150e+00(3.3e-03) & 1.191e+00(2.0e-02) & 1.291e+00(7.6e-03) \\
JES & 1.167e+00(6.6e-03) & 1.156e+00(3.2e-03) & 1.212e+00(1.3e-02) & 1.303e+00(4.5e-03) \\
PSL & 1.128e+00(3.9e-03) & 1.073e+00(4.5e-02) & 1.123e+00(2.2e-02) & 1.273e+00(7.6e-03) \\
BOFormer & 1.176e+00(3.1e-03) & 1.161e+00(3.6e-03) & 1.244e+00(6.3e-03) & 1.307e+00(3.3e-03) \\
TAMO & 1.176e+00(1.3e-03) & 1.164e+00(2.7e-03) & 1.244e+00(1.9e-02) & 1.309e+00(2.7e-03) \\
\hline
FoMEMO-EI & 1.177e+00(5.3e-03) & 1.173e+00(2.0e-03) & 1.261e+00(2.1e-02) & 1.315e+00(2.9e-03) \\
FoMEMO-UCB & 1.178e+00(7.3e-03) & \cellcolor{gray!10}\textbf{1.175e+00(1.9e-03)} & 1.262e+00(1.8e-02) & 1.316e+00(2.3e-03) \\
FoMEMO-UHVI & 1.177e+00(6.2e-03) & \cellcolor{gray!30}\textbf{1.176e+00(2.4e-03)} & \cellcolor{gray!10}\textbf{1.271e+00(1.6e-02)} & \cellcolor{gray!10}\textbf{1.318e+00(1.9e-03)} \\
FoMEMO-UR2I & \cellcolor{gray!30}\textbf{1.182e+00(2.6e-03)} & \cellcolor{gray!50}\textbf{1.179e+00(2.2e-03)} & \cellcolor{gray!50}\textbf{1.274e+00(1.2e-02)} & \cellcolor{gray!30}\textbf{1.320e+00(2.1e-03)} \\
\bottomrule[1pt]
\end{tabular}}
\end{table}

\section{Analysis of Runtime}
\label{apd: analysis of runtime}
\begin{table}[!h]
\centering
\caption{Comparison of average query time per iteration for all algorithms in sequential optimization ($q=1$).}\label{tab:query time}
\renewcommand{\arraystretch}{1.3}
\scalebox{0.7}{
\begin{tabular}{ccccc}
\toprule[1pt]
Algorithm & 2 Objectives & 3 Objectives & 4 Objectives & 6 Objectives  \\
\hline
MOEA/D-EGO & 38.6409 & 46.8428 & 49.9341 & 54.1983 \\
TSEMO & 5.6616 & 6.1670 & 5.8793 & 6.3524 \\
USEMO-EI & 8.1582 & 6.6809 & 7.2214 & 6.7517 \\
DGEMO & 36.2580 & 95.3770 & N/A & N/A \\
qParEGO & 11.9262 & 19.0253 & 85.5916 & 185.4954 \\    
qNEHVI & 10.1029 & 22.0746 & 62.9022 & 78.2812 \\        
JES & 10.3307 & 33.4365 & 41.6235 & 44.9701 \\
PSL & 0.0990 & 0.2065 & 0.1053 & 0.0916 \\
BOFormer & 0.6679 & 0.7483 & N/A & N/A \\
TAMO & 0.0173 & N/A & N/A & N/A \\
\hline
FoMEMO-EI & 2.5706 & 3.9841 & 3.1434 & 2.8501 \\      
FoMEMO-UCB & 1.7201 & 1.1666 & 1.1155 & 2.0801 \\
FoMEMO-UHVI & 0.6189 & 0.6094 & 0.6197 & 0.7432 \\
FoMEMO-UR2I & 1.6248 & 1.7077 & 1.3686 & 1.1716 \\
\bottomrule[1pt]
\end{tabular}}
\end{table}

We evaluate the computational efficiency in sequential optimization across 4 engineering design problems (four bar truss, speed reducer, car side impact design and water resource planning problems) with objective counts ranging from 2 to 6, as shown in Table \ref{tab:query time}. Beyond the inherent time advantage of bypassing model retraining, here we ensure a rigorous comparison by monitoring only the duration required to generate candidate solutions in each optimization iteration, excluding the time allocated to model retraining (where applicable) and true function evaluations. This allows us to isolate and highlight the efficiency gains provided by the foundation model’s in-context optimization capabilities. The reported query time is averaged from 10 independent runs with 100 iterations each. 

Due to the inference overhead of GP models and the computational cost of associated operations, such as the population-based optimization in MOEA/D-EGO, the intricate batch selection in DGEMO, and the expensive HV calculations in qNEHVI, the query time of conventional MOBO methods increases substantially with problem complexity. In contrast, benefiting from the fast in-context learning of our foundation model, FoMEMO achieves much lower query-generation cost when optimizing acquisition functions compared with conventional GP-based MOBO baselines, especially as the number of objectives increases. While BOFormer remains competitive in 2D and 3D objectives, it is still surpassed by FoMEMO-UHVI. We also observe that PSL and TAMO achieve very low query latency. This is expected, since they follow a direct-query paradigm: PSL directly learns a preference-to-solution mapping, while TAMO directly maps the optimization history to the next query through an end-to-end policy. Therefore, they avoid posterior inference and acquisition-function optimization at test time, giving them a natural advantage in query time. However, despite this query-time advantage, PSL and TAMO achieve lower optimization quality than FoMEMO in our experiments, as reflected in the optimization performance results.

Overall, these results suggest that FoMEMO provides a favorable balance between efficiency and optimization performance. Although direct-query models such as PSL and TAMO can be faster in terms of query time, FoMEMO remains highly efficient while retaining a posterior-based acquisition-function interface, which leads to stronger downstream optimization performance. Since true function evaluations often dominate model inference time in real-world expensive optimization problems, the low query-generation cost of FoMEMO remains practical, and its improved solution quality is especially valuable in real-world applications where sample efficiency is critical.

\section{Ablation Study}
\label{ablation_study}

\subsection{Ablation Study of Model Configurations}
\label{model_configuration_ablation}
In this subsection, we provide a more detailed analysis of the model training process. To further demonstrate the superiority of our selected model configuration, we compare the optimization results across different configurations. Here, “Base” refers to the model setup used throughout this paper, which employs an RBF kernel, an objective-aware regression head, and Tchebycheff aggregation. For ablation analysis, we retrain the model by altering one key component at a time while keeping the rest of the configurations unchanged. Specifically, we experiment with Mat\'ern-3/2 and Mat\'ern-5/2 kernels, a uniform regression head that is agnostic to the objective dimension, and a smoothed version of the Tchebycheff aggregation with a default smoothing parameter of 0.05 \cite{lin2024smooth}.

\begin{figure*}[htbp]
\centering
\includegraphics[scale=0.48]{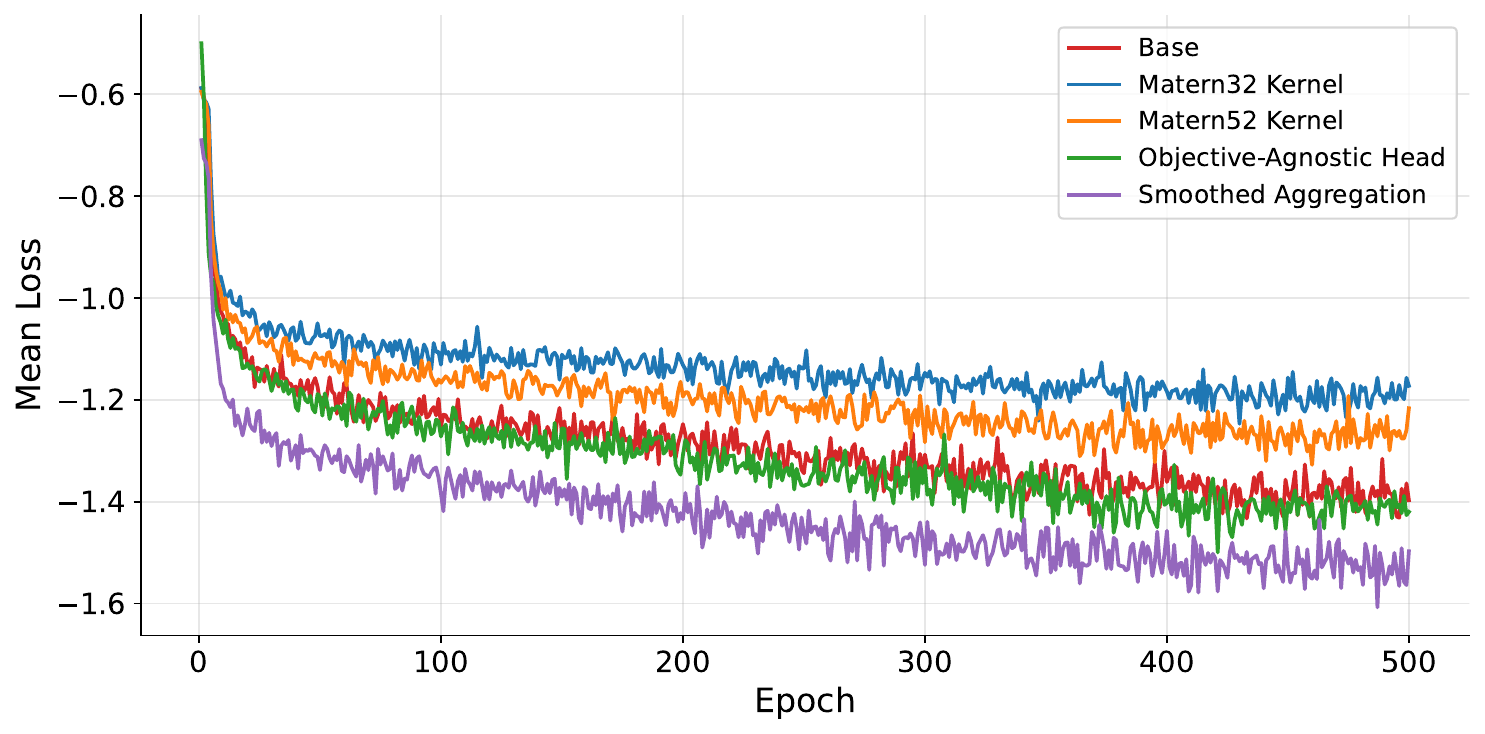}
\caption{The mean loss convergences with different model configurations.} 
\label{fig:model loss}
\end{figure*}

\paragraph{Training Loss} We monitor the convergences of the mean loss during training presented by model with different configurations, represented in Figure \ref{fig:model loss}. Despite minor fluctuations, each loss curve shows that it has essentially converged. We find that using the RBF kernel to generate data helps achieve a lower loss level, while whether or not a objective-aware regression head is used has little impact on the reduction of loss. Notably, using smoothed aggregation leads to faster convergence and lower loss, which is related to the numerical stability brought by smoothing operation \cite{lin2024smooth}, and thus facilitates gradient-based optimization of model parameters. However, there is a gap between the model pre-training phase and the subsequent downstream optimization process using the pre-trained models, and the performance of loss reduction cannot directly determine the optimization performance of the algorithm, which we will discuss further later.

\begin{table}[!h]
\centering
\caption{Averaged time usage per step during model training (in seconds).}\label{tab:model training time}
\renewcommand{\arraystretch}{1.5}
\scalebox{0.8}{
\begin{tabular}{cccc}
\toprule[1pt]
Model Configuration & Data Time & Step Time & Forward Time \\
\hline
Base & 0.0474 & 0.0323 & 0.0065 \\
Mat\'ern-3/2 Kernel & 0.0491 & 0.0325 & 0.0066 \\
Mat\'ern-5/2 Kernel & 0.0489 & 0.0326 & 0.0067 \\
Objective-Agnostic Head & 0.0518 & 0.0325 & 0.0100 \\
Smoothed Aggregation & 0.0588 & 0.0382 & 0.0067 \\
\bottomrule[1pt]
\end{tabular}}
\end{table}

\paragraph{Time Usage During Model Training} In our method, the data synthesis and model training are performed simultaneously. To evaluate the advantages of using synthetic data for pre-training and in-context optimization with a single forward pass, we record the detailed information on averaged time for each step utilized during training, as illustrated in Table \ref{tab:model training time}. Among them, ``data time'' refers to the time spent synthesizing data for each step, ``forward time'' indicates the duration spent on model inference to generate logits for loss calculation, and ``step time'' denotes the time taken to compute the loss and perform gradient backpropagation to optimize the model parameters. Statistics results indicate that for all model configurations, the model optimization and data synthesis account for the majority of pre-training time. Although a large amount of data is synthesized during training using our sampling mechanism, the time spent on data synthesis is comparable to the time dedicated to optimizing model parameters. This efficiency can be largely attributed to the effective computation of the decomposition-based aggregation targets from extensive synthetic functions, which does not impose an unacceptable burden on pre-training as would be the case with expensive experimental data from real-world problems. Furthermore, it is noteworthy that a single forward propagation through the neural network is quite fast, this efficiency helps explain the rapid in-context inference observed once the pre-training is completed.

\begin{table}[htbp]
\centering
\caption{Means of HV ($\Delta$ HV relative to base configuration) over 10 independent runs. The better mean result for each algorithm compared to the base configuration is marked in dark green.}\label{tab:model ablation exp}
\renewcommand{\arraystretch}{1.5}
\scalebox{0.5}{
\begin{tabular}{clllllll}
\toprule
Model Configuration & Algorithm & BC & ABD & Four Bar Truss & Gear Train & Materials & Mic \\
\midrule

\multirow{4}{*}{Base} & FoMEMO-EI & 8.066e-01 & 7.308e-01 & 8.667e-01 & 9.107e-01 & 1.173e+00 & 1.261e+00 \\
& FoMEMO-UCB & 8.075e-01 & 7.369e-01 & 8.722e-01 & 9.270e-01 & 1.175e+00 & 1.262e+00 \\
& FoMEMO-UHVI & 8.052e-01 & 7.430e-01 & 8.609e-01 & 8.573e-01 & 1.176e+00 & 1.271e+00 \\
& FoMEMO-UR2I & 8.016e-01 & 7.126e-01 & 8.741e-01 & 8.961e-01 & 1.179e+00 & 1.274e+00 \\
\cmidrule(l){2-8}

\multirow{4}{*}{Mat\'ern-3/2 Kernel} & FoMEMO-EI & 8.071e-01 \textcolor{mydarkgreen}{(+5.0e-04)} & 7.354e-01 \textcolor{mydarkgreen}{(+4.6e-03)} & 8.676e-01 \textcolor{mydarkgreen}{(+9.0e-04)} & 9.134e-01 \textcolor{mydarkgreen}{(+2.7e-03)} & 1.175e+00 \textcolor{mydarkgreen}{(+2.0e-03)} & 1.263e+00 \textcolor{mydarkgreen}{(+2.0e-03)} \\
& FoMEMO-UCB & 8.063e-01 (-1.2e-03) & 7.501e-01 \textcolor{mydarkgreen}{(+1.3e-02)} & 8.693e-01 (-2.9e-03) & 9.145e-01 (-1.3e-02) & 1.175e+00 (+0.0e+00) & 1.235e+00 (-2.7e-02) \\
& FoMEMO-UHVI & 7.841e-01 (-2.1e-02) & 7.340e-01 (-9.0e-03) & 8.560e-01 (-4.9e-03) & 8.456e-01 (-1.2e-02) & 1.175e+00 (-1.0e-03) & 1.261e+00 (-1.0e-02) \\
& FoMEMO-UR2I & 8.018e-01 \textcolor{mydarkgreen}{(+2.0e-04)} & 7.192e-01 \textcolor{mydarkgreen}{(+6.6e-03)} & 8.697e-01 (-4.4e-03) & 8.890e-01 (-7.1e-03) & 1.174e+00 (-5.0e-03) & 1.244e+00 (-3.0e-02) \\
\cmidrule(l){2-8}

\multirow{4}{*}{Mat\'ern-5/2 Kernel} & FoMEMO-EI & 8.074e-01 \textcolor{mydarkgreen}{(+8.0e-04)} & 7.430e-01 \textcolor{mydarkgreen}{(+1.2e-02)} & 8.687e-01 \textcolor{mydarkgreen}{(+2.0e-03)} & 9.090e-01 (-1.7e-03) & 1.175e+00 \textcolor{mydarkgreen}{(+2.0e-03)} & 1.254e+00 (-7.0e-03) \\
& FoMEMO-UCB & 8.083e-01 \textcolor{mydarkgreen}{(+8.0e-04)} & 7.495e-01 \textcolor{mydarkgreen}{(+1.3e-02)} & 8.736e-01 \textcolor{mydarkgreen}{(+1.4e-03)} & 9.156e-01 (-1.1e-02) & 1.177e+00 \textcolor{mydarkgreen}{(+2.0e-03)} & 1.253e+00 (-9.0e-03) \\
& FoMEMO-UHVI & 8.074e-01 \textcolor{mydarkgreen}{(+2.2e-03)} & 7.396e-01 (-3.4e-03) & 8.610e-01 \textcolor{mydarkgreen}{(+1.0e-04)} & 8.383e-01 (-1.9e-02) & 1.175e+00 (-1.0e-03) & 1.267e+00 (-4.0e-03) \\
& FoMEMO-UR2I & 7.975e-01 (-4.1e-03) & 7.178e-01 \textcolor{mydarkgreen}{(+5.2e-03)} & 8.753e-01 \textcolor{mydarkgreen}{(+1.2e-03)} & 8.658e-01 (-3.0e-02) & 1.178e+00 (-1.0e-03) & 1.255e+00 (-1.9e-02) \\
\cmidrule(l){2-8}

\multirow{4}{*}{\shortstack[c]{Objective-independent\\Head}} & FoMEMO-EI & 8.073e-01 \textcolor{mydarkgreen}{(+7.0e-04)} & 7.192e-01 (-1.2e-02) & 8.666e-01 (-1.0e-04) & 9.105e-01 (-2.0e-04) & 1.173e+00 (+0.0e+00) & 1.261e+00 (+0.0e+00) \\
& FoMEMO-UCB & 8.085e-01 \textcolor{mydarkgreen}{(+1.0e-03)} & 7.305e-01 (-6.4e-03) & 8.721e-01 (-1.0e-04) & 9.221e-01 (-4.9e-03) & 1.177e+00 \textcolor{mydarkgreen}{(+2.0e-03)} & 1.263e+00 \textcolor{mydarkgreen}{(+1.0e-03)} \\
& FoMEMO-UHVI & 8.070e-01 \textcolor{mydarkgreen}{(+1.8e-03)} & 7.349e-01 (-8.1e-03) & 8.639e-01 \textcolor{mydarkgreen}{(+3.0e-03)} & 8.506e-01 (-6.7e-03) & 1.174e+00 (-2.0e-03) & 1.267e+00 (-4.0e-03) \\
& FoMEMO-UR2I & 8.069e-01 \textcolor{mydarkgreen}{(+5.3e-03)} & 7.158e-01 \textcolor{mydarkgreen}{(+3.2e-03)} & 8.769e-01 \textcolor{mydarkgreen}{(+2.8e-03)} & 8.838e-01 (-1.2e-02) & 1.176e+00 (-3.0e-03) & 1.262e+00 (-1.2e-02) \\
\cmidrule(l){2-8}

\multirow{4}{*}{Smooth Aggregation} & FoMEMO-EI & 7.995e-01 (-7.1e-03) & 7.407e-01 \textcolor{mydarkgreen}{(+9.9e-03)} & 8.629e-01 (-3.8e-03) & 9.064e-01 (-4.3e-03) & 1.176e+00 \textcolor{mydarkgreen}{(+3.0e-03)} & 1.271e+00 \textcolor{mydarkgreen}{(+1.0e-02)} \\
& FoMEMO-UCB & 8.027e-01 (-4.8e-03) & 7.397e-01 \textcolor{mydarkgreen}{(+2.8e-03)} & 8.683e-01 (-3.9e-03) & 9.279e-01 \textcolor{mydarkgreen}{(+9.0e-04)} & 1.177e+00 \textcolor{mydarkgreen}{(+2.0e-03)} & 1.269e+00 \textcolor{mydarkgreen}{(+7.0e-03)} \\
& FoMEMO-UHVI & 7.977e-01 (-7.5e-03) & 7.280e-01 (-1.5e-02) & 8.577e-01 (-3.2e-03) & 8.329e-01 (-2.4e-02) & 1.174e+00 (-2.0e-03) & 1.266e+00 (-5.0e-03) \\
& FoMEMO-UR2I & 7.903e-01 (-1.1e-02) & 6.897e-01 (-2.3e-02) & 8.669e-01 (-7.2e-03) & 8.857e-01 (-1.0e-02) & 1.176e+00 (-3.0e-03) & 1.267e+00 (-7.0e-03) \\

\bottomrule
\end{tabular}}
\end{table}

\paragraph{Optimization Performance} We further compare the optimization performance of all proposed acquisition functions with different model configurations on 6 representative problems, as shown in Table \ref{tab:model ablation exp}. For each algorithm with different model configurations, if it outperforms the corresponding algorithm in base configuration, we mark it in dark green to visualize the impact of different model configurations on the final optimization performance. The results show that our chosen model configuration performs better in most cases. We find that using Mat\'ern-3/2 and Mat\'ern-5/2 kernels yield better results especially for synthetic problems involving Branin functions. The reason may lie that the RBF kernel, being infinitely differentiable, imposes a strong smoothness prior that hinders its ability to capture the localized transitions and high-curvature regions between the peaks and valleys of the Branin landscape. In contrast, the Mat\'ern class of kernels relaxes these smoothness constraints, which balances global trend modeling and local flexibility required to navigate the distinctive landscape of the Branin function. In addition, we observe that using the objective-agnostic head yields inferior results compared to the base configuration in most cases. This demonstrates that while loss gaps may be marginal during training, the adaptive selection of dimension-specific regressors during inference is pivotal for enhancing final optimization outcomes. Finally, empirical results indicate that the configuration with smoothed aggregation generally underperforms the baseline. This disparity likely stems from the fact that while smoothing facilitates model training, the approximation errors introduced by smoothing operation may bias the accurate fitting of the optimal region in the exact aggregation function \cite{lin2024smooth}, further misleading the precise identification of the global optimum in the sub-problem during the optimization process.

\subsection{Ablation Study of Number of Preferences in UHVI and UR2I}
\label{apd:UHVI pref ablation}

\begin{figure}[htbp]
\centering
\includegraphics[scale=0.27]{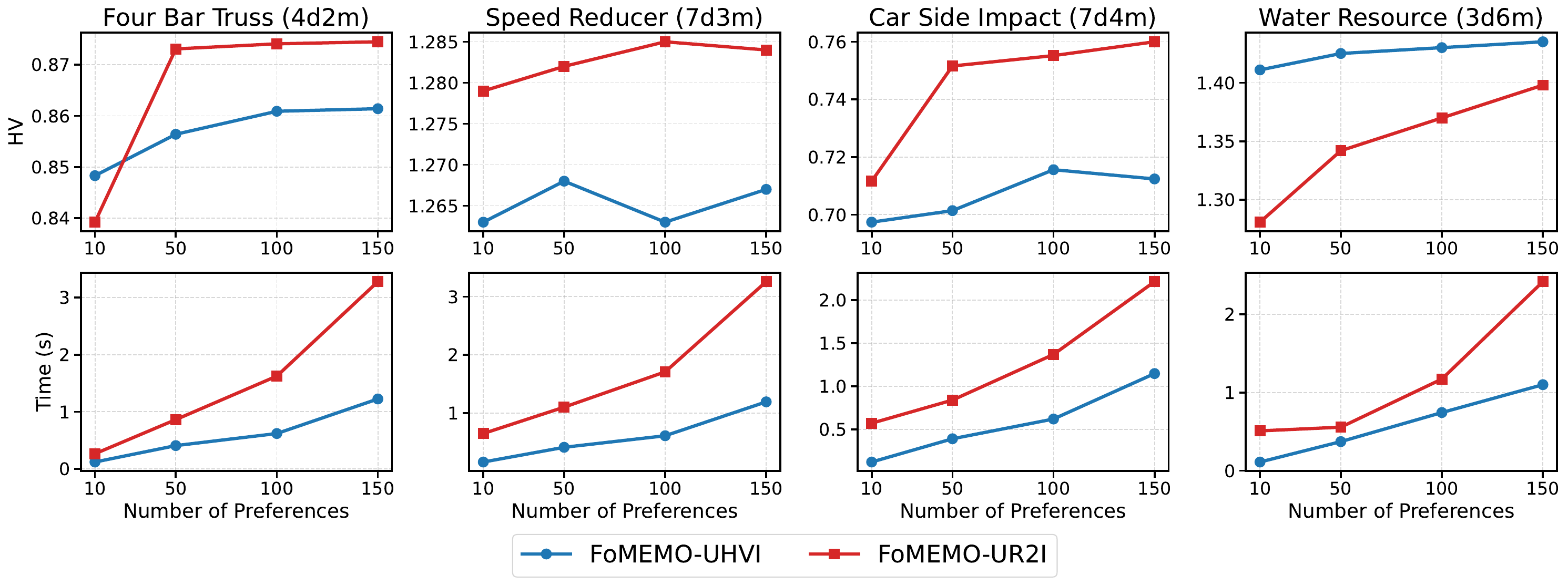}
\caption{The hypervolume metrics and averaged query time with different number of preferences.}
\label{fig:preference free af ablation}
\end{figure}

In this subsection, we investigate the effect of the number of sampled preference vectors on the preference-free acquisition functions UHVI and UR2I. Specifically, we evaluate 10, 50, 100, and 150 sampled preference vectors on problems with different numbers of objectives, namely 2, 3, 4, and 6 objectives. The results are shown in Figure \ref{fig:preference free af ablation}. In most cases, using more sampled preference vectors improves the optimization metric, but also increases the average query-generation time. To balance optimization performance and computational cost, we use 100 sampled preference vectors for computing UHVI and UR2I throughout this paper.

\subsection{Ablation Study of PFN-Based Methods}
\label{pfn_based_methods_ablation}

\begin{table}[!h]
\centering
\caption{Predictive accuracy and uncertainty calibration comparison.}
\label{tab:uncertainty-calibration}
\renewcommand{\arraystretch}{1.5}
\scalebox{0.7}{
\begin{tabular}{llcccccc}
\toprule[1pt]
Problem & Method & RMSE & MAE & $R^2$ & NLL & Coverage@95\% & MPIW \\
\hline
\multirow{5}{*}{\shortstack[l]{Four Bar Truss\\(2 Objectives)}}
& FoMEMO & \cellcolor{gray!50}\textbf{0.0085} & \cellcolor{gray!50}\textbf{0.0057} & \cellcolor{gray!50}\textbf{0.9925} & \cellcolor{gray!50}\textbf{-3.5560} & \cellcolor{gray!50}\textbf{0.9486} & 0.0316 \\
& PFNs & 0.0334 & 0.0262 & 0.7973 & -1.5384 & 1.0000 & 0.3174 \\
& GP-RBF & 0.0114 & 0.0076 & 0.9709 & -3.0026 & 0.7957 & 0.0241 \\
& GP-Mat\'ern-3/2 & 0.0103 & 0.0065 & 0.9854 & -3.1493 & 0.9700 & 0.0385 \\
& GP-Mat\'ern-5/2 & 0.0091 & 0.0059 & 0.9857 & -3.4093 & 0.9257 & 0.0277 \\
\hline
\multirow{5}{*}{\shortstack[l]{Speed Reducer\\(3 Objectives)}}
& FoMEMO & 0.0287 & 0.0202 & 0.9214 & -2.4440 & 0.9643 & 0.1115 \\
& PFNs & 0.0357 & 0.0258 & 0.8251 & -1.5385 & 0.9871 & 0.3033 \\
& GP-RBF & 0.0123 & 0.0086 & 0.9755 & -2.0039 & 0.7629 & 0.0263 \\
& GP-Mat\'ern-3/2 & \cellcolor{gray!50}\textbf{0.0087} & \cellcolor{gray!50}\textbf{0.0062} & \cellcolor{gray!50}\textbf{0.9873} & \cellcolor{gray!50}\textbf{-3.5847} & \cellcolor{gray!50}\textbf{0.9400} & 0.0307 \\
& GP-Mat\'ern-5/2 & 0.0102 & 0.0074 & 0.9829 & -3.3175 & 0.8886 & 0.0283 \\
\hline
\multirow{5}{*}{\shortstack[l]{Car Side Impact\\(4 Objectives)}}
& FoMEMO & \cellcolor{gray!50}\textbf{0.0126} & \cellcolor{gray!50}\textbf{0.0088} & \cellcolor{gray!50}\textbf{0.9392} & \cellcolor{gray!50}\textbf{-3.0036} & 0.9843 & 0.0686 \\
& PFNs & 0.0372 & 0.0282 & 0.4627 & -1.5470 & 0.9943 & 0.3043 \\
& GP-RBF & 0.0156 & 0.0109 & 0.9068 & -2.5260 & 0.8657 & 0.0436 \\
& GP-Mat\'ern-3/2 & 0.0132 & 0.0089 & 0.9276 & -2.9158 & \cellcolor{gray!50}\textbf{0.9471} & 0.0485 \\
& GP-Mat\'ern-5/2 & 0.0138 & 0.0093 & 0.9235 & -2.8221 & 0.9157 & 0.0459 \\
\hline
\multirow{5}{*}{\shortstack[l]{Water Resource\\(6 Objectives)}}
& FoMEMO & \cellcolor{gray!50}\textbf{0.0194} & \cellcolor{gray!50}\textbf{0.0071} & \cellcolor{gray!50}\textbf{0.8135} & \cellcolor{gray!50}\textbf{-4.2586} & \cellcolor{gray!50}\textbf{0.9400} & 0.0319 \\
& PFNs & 0.0320 & 0.0173 & 0.5916 & -1.5985 & 0.9900 & 0.2920 \\
& GP-RBF & 0.0233 & 0.0096 & 0.7343 & 0.2291 & 0.7986 & 0.0391 \\
& GP-Mat\'ern-3/2 & 0.0237 & 0.0078 & 0.7005 & -2.2236 & 0.9014 & 0.0442 \\
& GP-Mat\'ern-5/2 & 0.0260 & 0.0089 & 0.6142 & -1.4038 & 0.8614 & 0.0376 \\
\bottomrule[1pt]
\end{tabular}}
\end{table}

\paragraph{Predictive Accuracy and Uncertainty Calibration} We first compare FoMEMO with the official pretrained PFN model \cite{muller2023pfns4bo} to provide a more comprehensive assessment of predictive accuracy and uncertainty calibration for aggregated posteriors. We add negative log-likelihood (NLL), empirical 95\% coverage, and mean prediction interval width (MPIW) on four OOD real-world problems covering 2, 3, 4, and 6 objectives, as shown in Table \ref{tab:uncertainty-calibration}. In each of five independent runs, all methods shared the same 200 Sobol points and the same randomly generated 30\%/70\% split (60 observed and 140 held-out points). Compared with PFNs, FoMEMO is better calibrated on all four problems. PFNs' 98.71\%-100\% coverage deviates further from 95\% and relies on much wider intervals (MPIW 0.2920-0.3174 versus 0.0316-0.1115). Compared with exact GP variants, FoMEMO obtains the best RMSE, MAE, $R^2$ and NLL on three of the four problems. Its empirical coverage of 94.00\%-98.43\% remains close to the nominal 95\% and competitive with the exact GPs without per-task fitting. These results support both predictive accuracy and a favorable calibration balance of our method.

\paragraph{Objective-wise PFNs4BO Baseline} We then compare FoMEMO with the objective-wise PFNs4BO baselines in terms of downstream optimization performance and query efficiency. The official pretrained PFNs4BO model \cite{muller2023pfns4bo} predicts each objective independently, after which the objective-wise predictions are combined through scalarization or hypervolume improvement. Specifically, PFNs4BO-UCB applies random Tchebycheff scalarization to the objective-wise UCBs, while PFNs4BO-UHVI computes hypervolume improvement from the objective-wise UCB vector. Compared with FoMEMO-UCB and FoMEMO-UHVI, this provides matched acquisition-family comparisons between post-hoc aggregation of objective-wise PFN predictions and direct aggregated posterior learning, as shown in Tables \ref{tab:pfns4bo-hv} and \ref{tab:pfns4bo-query-time}. Across six synthetic and real-world tasks, each FoMEMO variant outperforms its matched PFNs4BO counterpart in final hypervolume on every task. FoMEMO additionally achieves a lower average query time in six of the eight matched comparisons across different objectives. These consistent gains show that preference-wise aggregated posterior learning provides a clear advantage over the simpler objective-wise PFN construction within the in-context multi-objective optimization paradigm.

\begin{table}[!h]
\centering
\caption{Final hypervolume comparison against PFNs4BO baselines.}
\label{tab:pfns4bo-hv}
\renewcommand{\arraystretch}{1.5}
\scalebox{0.7}{
\begin{tabular}{ccccccc}
\toprule[1pt]
Method & BC & ABD & Four Bar Truss & Gear Train & Materials & Mic \\
\hline
PFNs4BO-UCB & 8.049e-1(1.4e-3) & 6.993e-1(1.2e-2) & 8.465e-1(4.3e-3) & 8.163e-1(4.8e-2) & 1.163e+0(7.0e-3) & 1.244e+0(1.3e-2) \\
PFNs4BO-UHVI & 7.734e-1(6.5e-3) & 5.644e-1(4.1e-2) & 8.023e-1(7.8e-3) & 6.612e-1(1.3e-1) & 1.149e+0(1.4e-2) & 1.217e+0(1.7e-2) \\
\hline
FoMEMO-UCB & \cellcolor{gray!50}\textbf{8.075e-1(5.8e-4)} & 7.369e-1(8.5e-3) & \cellcolor{gray!50}\textbf{8.722e-1(1.2e-3)} & \cellcolor{gray!50}\textbf{9.270e-1(4.5e-3)} & 1.175e+0(1.9e-3) & 1.262e+0(1.8e-2) \\
FoMEMO-UHVI & 8.052e-1(4.9e-4) & \cellcolor{gray!50}\textbf{7.430e-1(8.4e-3)} & 8.609e-1(1.3e-3) & 8.573e-1(9.6e-3) & \cellcolor{gray!50}\textbf{1.176e+0(2.4e-3)} & \cellcolor{gray!50}\textbf{1.271e+0(1.6e-2)} \\
\bottomrule[1pt]
\end{tabular}}
\end{table}

\begin{table}[!h]
\centering
\caption{Average query-time comparison against PFNs4BO baselines.}
\label{tab:pfns4bo-query-time}
\renewcommand{\arraystretch}{1.5}
\scalebox{0.7}{
\begin{tabular}{ccccc}
\toprule[1pt]
Method & 2 Objectives & 3 Objectives & 4 Objectives & 6 Objectives \\
\hline
PFNs4BO-UCB & 3.2040 & 0.9760 & 3.0006 & 1.5506 \\
PFNs4BO-UHVI & 0.8773 & 1.9742 & 1.3186 & 0.8821 \\
\hline
FoMEMO-UCB & 1.7201 & 1.1666 & 1.1155 & 2.0801 \\
FoMEMO-UHVI & 0.6189 & 0.6094 & 0.6197 & 0.7432 \\
\bottomrule[1pt]
\end{tabular}}
\end{table}

\paragraph{Insights into Aggregated Posterior Learning} These comparisons provide further insights into the benefits of directly learning preference-wise aggregated posteriors. Since the Tchebycheff max operation generally induces an unknown and non-Gaussian aggregated posterior, both Gaussian aggregation approximation, as in MOEA/D-EGO, and GP modeling of individual scalarized subproblems, as in qParEGO, will introduce approximation errors and compromise predictive fidelity and uncertainty quantification, thereby distorting the acquisition landscape and ultimately degrading candidate quality. FoMEMO instead learns the unknown aggregated posterior directly without sacrificing predictive fidelity or uncertainty quantification. The predictive-calibration results in Table \ref{tab:uncertainty-calibration} show that the learned aggregated posterior provides accurate predictions and reliable uncertainty estimates. The objective-wise PFNs4BO comparisons in Table \ref{tab:pfns4bo-hv} further show that directly learning this posterior leads to better downstream optimization than predicting each objective separately and aggregating the predictions afterward. Together, these results help isolate the algorithmic benefit of preference-wise aggregated posterior learning from the efficiency gain achieved by eliminating online fitting with a pretrained PFN model.

\section{Role of Aggregated Posterior Learning}
\label{app:aggregated_posterior_learning}

In this section, we first state our main contribution, identify the MOBO bottlenecks it addresses, explain how aggregated posterior learning as the key component supports this contribution, and identify the sources of the observed performance and wall-time improvements.

\subsection{Main Contribution Statement}
\label{app:main_contribution_statement}

Our main contribution is a scalable, decomposition-based pre-training paradigm that reformulates MOBO as preference-wise aggregated posterior learning, which is motivated by the need for broad and efficient adaptation across diverse emerging real-world applications. This paradigm addresses key limitations of existing MOBO approaches by simultaneously achieving scalable pre-training, amortized surrogate modeling, high-fidelity posterior prediction, and efficient acquisition optimization.

\subsection{MOBO Bottlenecks}
\label{app:mobo_bottlenecks}

\paragraph{Repeated Surrogate Modeling} Conventional GP-based MOBO methods require constructing task-specific surrogates from scratch for every new problem and repeatedly updating them during optimization, imposing both a substantial modeling burden and recurring online computational overhead.

\paragraph{Modeling Fidelity Limitations} Existing MOBO methods often rely on restrictive assumptions in aggregated posterior modeling. Since the Tchebycheff max operation induces an unknown and generally non-Gaussian aggregated posterior, whether using a Gaussian aggregation approximation, as in MOEA/D-EGO, or modeling individual scalarized subproblems with GPs, as in qParEGO, will introduce approximation errors and compromise predictive fidelity and uncertainty quantification, thereby distorting the acquisition landscape and ultimately degrading candidate quality.

\paragraph{Acquisition Burden} In conventional MOBO, acquisition optimization repeatedly evaluates the posterior during candidate search. The costs of model updating and inference therefore propagate into the acquisition stage and amplify its computational burden, especially when identifying the unknown Pareto front is challenging or candidate search involves computationally expensive procedures such as hypervolume computation and population-based optimization.

\paragraph{Training Scalability} Recent amortized methods such as BOFormer and TAMO face limited training scalability because they rely on hypervolume-oriented acquisition or query-policy learning. As shown in Table~\ref{tab:metric computation time}, these training objectives become computationally demanding when scaled to massive synthetic data generation, limiting the diversity of the pre-training distribution and robust generalization to unseen scenarios.

\subsection{Role of Aggregated Posterior Learning in Addressing MOBO Bottlenecks}
\label{app:aggregated_posterior_role}

Our core contribution is the scalable pre-training paradigm. Preference-wise aggregated posterior learning is the central mechanism that realizes this paradigm, while the acquisition functions convert the learned posterior into practical MOBO sampling functionality. The three elements therefore correspond to the overall paradigm, its posterior-learning mechanism, and its downstream optimization components. We next explain how aggregated posterior learning serves as the key component that supports this paradigm and addresses the bottlenecks in existing MOBO approaches identified above.

\paragraph{Amortized Surrogate Modeling} Learning a shared posterior family within a single model amortizes surrogate modeling across tasks and preferences, eliminating repeated per-problem and per-subproblem fitting required by traditional MOBO.

\paragraph{Enhanced Posterior Fidelity} Directly learning the unknown aggregated posterior avoids the modeling compromises introduced by Gaussian approximation or post-hoc aggregation of independently predicted objectives, without sacrificing predictive fidelity or uncertainty quantification. The predictive-calibration results in Table~\ref{tab:uncertainty-calibration} show that the learned aggregated posterior provides accurate predictions and reliable uncertainty estimates.

\paragraph{Efficient Acquisition Optimization} The resulting aggregated posterior offers an efficient interface for repeated acquisition queries, thereby alleviating the model-side computational burden of acquisition optimization.

\paragraph{Scalable Pre-training} By introducing preference-wise aggregated posterior learning, supervised target generation within the proposed paradigm becomes scalable across objective dimensions and dataset sizes. Unlike the hypervolume-oriented training objectives used by BOFormer and TAMO, Tchebycheff targets can be computed efficiently, as shown in Table~\ref{tab:metric computation time}, enabling pre-training on a diverse corpus of more than 130 million synthetic datasets and thereby supporting robust generalization to unseen complex problems.

\subsection{Attribution of Performance and Wall-Time Improvements}
\label{app:improvement_attribution}

We summarize the ablation evidence below to identify which component of the method accounts for the observed performance gains and wall-time reductions.

\paragraph{Predictive Performance} Table~\ref{tab:uncertainty-calibration} compares alternative approaches to aggregated posterior modeling under identical evaluation protocols. FoMEMO obtains the best point metrics and NLL on three of four problems and maintains coverage close to the nominal 95\% level. These results support attributing the gains in predictive fidelity and calibration to directly learning the unknown aggregated posterior, rather than relying on task-specific GP surrogates or using a pretrained PFN as a direct alternative.

\paragraph{Optimization Performance} Table~\ref{tab:pfns4bo-hv} holds matched acquisition variants and the evaluation protocols fixed. Each FoMEMO variant outperforms its matched PFNs4BO counterpart on all six tasks, showing that directly learning this posterior leads to better downstream optimization than predicting each objective separately and aggregating the predictions afterward, supporting attribution of the matched downstream gains to the aggregated posterior modeling rather than from eliminating online training and the acquisition family.

\paragraph{Training-Data Generation Time} Table~\ref{tab:metric computation time} compares the training-data generation times of Tchebycheff aggregation and hypervolume computation under the same dataset length and computational protocol. The hypervolume-to-Tchebycheff runtime ratio increases from $6.79\times$ at two objectives to approximately $4.6\times10^5$ at six objectives. This controlled comparison attributes training scalability to the computational efficiency of Tchebycheff aggregation, which makes pre-training on large diverse synthetic datasets practical.

\paragraph{Query Time} In Table~\ref{tab:pfns4bo-query-time}, neither method in the matched PFNs4BO comparison performs per-task refitting. Although FoMEMO has lower query time in six of the eight comparisons, the differences are smaller than those observed against most conventional MOBO methods, as shown in Table~\ref{tab:query time}. The smaller query-time gap against PFNs4BO suggests that the efficiency advantage over conventional MOBO is associated in part with amortized posterior inference, while the remaining differences may reflect the posterior representation and acquisition-query interface.

Together, the ablation evidence above shows that preference-wise aggregated posterior learning provides the key mechanism for accurate posterior modeling, improved downstream optimization, scalable training-data generation, and efficient amortized inference.

\section{Analysis of Method Scalability}
\label{apd:method_scalability}
\subsection{Scalability Within Pre-training Dimensions}
\label{app:scalability_within_pretraining_dimensions}

\begin{table}[htbp]
\centering
\caption{Means (stds) of hypervolume metrics ($\uparrow$) on 3 engineering design problems with many objectives and a multi-modal WFG problem with complex search space over 10 independent runs. The best, second-best, and third-best mean results for each problem are highlighted with dark gray, gray, and light gray backgrounds, respectively.}\label{tab:high dim exp}
\renewcommand{\arraystretch}{1.5}
\scalebox{0.7}{
\begin{tabular}{ccccc}
\toprule[1pt]
Algorithm & Car Side Impact (7d4m) & Conceptual Marine (6d4m) & Water Resource (3d6m) & WFG4 (30d3m) \\
\hline
MOEA/D-EGO & 7.039e-01(5.7e-03) & 5.915e-01(2.4e-02) & 1.412e+00(1.3e-02) & 2.754e-01(5.3e-03) \\
TSEMO & 6.942e-01(7.9e-03) & 7.358e-01(7.9e-03) & 1.198e+00(5.7e-02) & 2.869e-01(4.3e-03) \\
USEMO-EI & 7.151e-01(7.2e-03) & 6.750e-01(1.1e-02) & \cellcolor{gray!10}\textbf{1.433e+00(9.4e-03)} & 2.842e-01(5.3e-03) \\
DGEMO & N/A & N/A & N/A & 2.792e-01(4.6e-03) \\
qNEHVI & 5.900e-01(7.0e-03) & 6.472e-01(2.2e-02) & 1.342e+00(4.8e-02) & \cellcolor{gray!10}\textbf{3.276e-01(2.2e-02)} \\
qParEGO & 5.740e-01(2.3e-02) & 3.887e-01(2.6e-02) & 1.404e+00(3.7e-02) & 3.213e-01(1.5e-02) \\
JES & 5.629e-01(1.7e-02) & 4.628e-01(2.9e-02) & 1.296e+00(2.6e-02) & \cellcolor{gray!30}\textbf{3.314e-01(7.5e-03)} \\
PSL & 5.415e-01(2.3e-02) & 3.237e-01(5.7e-02) & 1.148e+00(5.0e-02) & 1.668e-01(1.6e-02) \\
BOFormer & N/A & N/A & N/A & 3.086e-01(5.2e-03) \\
TAMO & N/A & N/A & N/A & N/A \\
\hline
FoMEMO-EI & \cellcolor{gray!10}\textbf{7.538e-01(7.4e-03)} & \cellcolor{gray!10}\textbf{7.505e-01(1.9e-02)} & \cellcolor{gray!30}\textbf{1.470e+00(4.7e-03)} & 3.159e-01(4.3e-03) \\
FoMEMO-UCB & \cellcolor{gray!50}\textbf{7.739e-01(8.1e-03)} & \cellcolor{gray!50}\textbf{7.814e-01(1.5e-02)} & \cellcolor{gray!50}\textbf{1.475e+00(3.9e-03)} & 3.150e-01(4.4e-03) \\
FoMEMO-UHVI & 7.156e-01(4.7e-03) & 6.012e-01(1.6e-02) & 1.430e+00(1.9e-02) & 3.173e-01(7.8e-03) \\
FoMEMO-UR2I & \cellcolor{gray!30}\textbf{7.552e-01(5.1e-03)} & \cellcolor{gray!30}\textbf{7.587e-01(3.8e-03)} & 1.370e+00(2.0e-02) & \cellcolor{gray!50}\textbf{3.397e-01(1.4e-02)} \\
\bottomrule[1pt]
\end{tabular}}
\end{table}

In this subsection, we further evaluate the dimensional scalability of our method, including both objective and feature dimensions. Firstly, we consider 3 more challenging real-world engineering design problems with higher objective dimensions, including car side impact design and conceptual marine design problems, both with 4 objectives, as well as water resource planning problem, which has 6 objectives \cite{tanabe2020easy}. The experimental results are shown in Table \ref{tab:high dim exp}. We find that our method demonstrates superior scalability and maintains robust performance in many-objective optimization scenarios. A primary factor contributing to this efficacy of our approach is its ability to directly predict the exact aggregated posterior as a universal basis, which avoids the performance-degrading compromises often found in existing methods. In contrast, traditional methods suffer from significant limitations as the objective dimensionality increases: qNEHVI faces the computational burden of non-dominated partitioning and hypervolume calculations, which become computationally intractable in high-dimensional spaces. MOEA/D-EGO introduces approximation errors due to the non-Gaussian nature of the max operation when using Gaussian distribution to estimate the unknown aggregation distribution in high-dimensional objective spaces. By bypassing these approximations, our method preserves modeling fidelity without sacrificing efficiency.

To examine the scalability of our approach across varying feature dimensions, we employ the WFG4 benchmark \cite{huband2006review}, which allows for flexible expansion of the decision space. Characterized by its pronounced multi-modality, WFG4 presents a complex landscape that rigorously evaluates an algorithm's ability to maintain global exploration and circumvent premature convergence. For this evaluation, we consider a 30-dimensional instance as a representative case. This setup aligns with the maximum dimensionality encountered during our current training phase, serving as a critical testbed to verify the model’s generalization robustness at the boundary of its training scope. Experimental results show that FoMEMO-UR2I consistently maintains superior performance even in this challenging scenario.

\subsection{Scalability Beyond Pre-training Dimensions}
\label{app:scalability_beyond_pretraining_dimensions}
To assess scalability beyond the pre-training ranges, we conduct additional out-of-distribution evaluations on 40-dimensional WFG4, 9-objective Car Cab Design, and Gear Train Design with a total of 310 evaluations, exceeding the respective pre-training ranges of 30 variables, 6 objectives, and 256 observations. For the 9-objective Car Cab Design problem, we use IGD+ performance metric (lower is better) because exact HV computation becomes computationally expensive. Tables~\ref{tab:beyond_pretraining_performance} and~\ref{tab:beyond_pretraining_query_time} report the final performance and average query time, respectively. FoMEMO remains effective across all three extrapolation settings, achieving the best final HV on WFG4 and Gear Train Design and competitive IGD+ on Car Cab Design. Its average query time is substantially lower on the 40-dimensional and 9-objective tests and remains manageable at 310 evaluations.

\begin{table}[htbp]
\centering
\caption{Means (stds) of performance metrics across settings beyond the pre-training ranges over 10 independent runs. The best, second-best, and third-best mean results for each problem are highlighted with dark gray, gray, and light gray backgrounds, respectively.}
\label{tab:beyond_pretraining_performance}
\renewcommand{\arraystretch}{1.5}
\scalebox{0.7}{
\begin{tabular}{cccc}
\toprule[1pt]
Method & WFG4 (40d3m) & Car Cab (7d9m) & Gear Train (310 evals.) \\
\hline
MOEA/D-EGO & 2.879e-1(5.0e-3) & \cellcolor{gray!30}\textbf{1.575e-1(3.4e-3)} & 6.162e-1(1.2e-1) \\
TSEMO & 2.959e-1(6.0e-3) & N/A & 8.619e-1(2.7e-2) \\
USeMO-EI & 2.984e-1(2.0e-3) & \cellcolor{gray!50}\textbf{1.466e-1(2.8e-3)} & 7.962e-1(2.7e-2) \\
qParEGO & \cellcolor{gray!10}\textbf{3.164e-1(1.3e-2)} & N/A & 8.488e-1(2.0e-2) \\
qNEHVI & 3.076e-1(7.1e-3) & N/A & \cellcolor{gray!10}\textbf{8.915e-1(1.1e-2)} \\
JES & 3.155e-1(8.7e-3) & N/A & 6.878e-1(8.1e-2) \\
\hline
FoMEMO-EI & \cellcolor{gray!30}\textbf{3.252e-1(5.6e-3)} & \cellcolor{gray!10}\textbf{1.662e-1(4.1e-3)} & \cellcolor{gray!30}\textbf{9.214e-1(4.9e-3)} \\
FoMEMO-UCB & 3.118e-1(3.5e-3) & 1.940e-1(9.1e-3) & \cellcolor{gray!50}\textbf{9.392e-1(1.5e-3)} \\
FoMEMO-UHVI & \cellcolor{gray!50}\textbf{3.274e-1(4.1e-3)} & 2.000e-1(8.9e-3) & 8.804e-1(7.2e-3) \\
FoMEMO-UR2I & 3.081e-1(4.8e-3) & 1.899e-1(8.5e-3) & 8.544e-1(2.2e-2) \\
\bottomrule[1pt]
\end{tabular}}
\end{table}

\begin{table}[htbp]
\centering
\caption{Comparison of average query time per iteration for all algorithms across settings beyond the pre-training ranges.}
\label{tab:beyond_pretraining_query_time}
\renewcommand{\arraystretch}{1.5}
\scalebox{0.7}{
\begin{tabular}{cccc}
\toprule[1pt]
Method & WFG4 (40d3m) & Car Cab (7d9m) & Gear Train (310 evals.) \\
\hline
MOEA/D-EGO & 37.2543 & 51.6466 & 31.8981 \\
TSEMO & 7.1097 & N/A & 2.6140 \\
USeMO-EI & 7.5294 & 7.9049 & 3.0156 \\
qParEGO & 27.5029 & N/A & 721.5189 \\
qNEHVI & 15.0819 & N/A & 414.0605 \\
JES & 8.8853 & N/A & 212.3084 \\
\hline
FoMEMO-EI & 2.0929 & 0.6106 & 10.9862 \\
FoMEMO-UCB & 0.4086 & 0.5665 & 8.7236 \\
FoMEMO-UHVI & 1.2500 & 0.3601 & 15.2566 \\
FoMEMO-UR2I & 2.1389 & 1.5230 & 66.4474 \\
\bottomrule[1pt]
\end{tabular}}
\end{table}

\subsection{Scalability in Parallel Optimization}
\label{app:scalability_in_parallel_optimization}

In this subsection, we provide supplementary experimental results to further validate the parallel scalability of our approach. Tables \ref{tab:q5 exp} and \ref{tab:q10 exp} report the performance across all algorithms with batch sizes of $q=5$ and $q=10$ per iteration, respectively. To ensure a rigorous and consistent comparison, a total evaluation budget of 100 samples is allocated for each configuration, excluding the initial design points. Our findings reveal that even with a straightforward batch generation strategy, our method consistently outperforms methods such as TSEMO and DGEMO in most cases, despite their reliance on sophisticated batch selection heuristics. Notably, among our proposed variants, the two preference-based acquisition functions yield the most superior results. This performance gap underscores the advantage of leveraging a foundation model pre-trained on diverse synthetic landscapes. Furthermore, it demonstrates that modeling aggregated posteriors parameterized by preferences is inherently amenable to parallelization, providing a more robust and natural framework for multi-objective optimization without focusing on developing more sophisticated batch-selection strategies.

\begin{table}[htbp]
\centering
\caption{Means (stds) of hypervolume metrics ($\uparrow$) in parallel optimization ($q=5$) on representative problems over 10 independent runs. The best, second-best, and third-best mean results for each problem are highlighted with dark gray, gray, and light gray backgrounds, respectively.}\label{tab:q5 exp}
\renewcommand{\arraystretch}{1.5}
\scalebox{0.63}{
\begin{tabular}{ccccccc}
\toprule[1pt]
Algorithm & BC & ABD & Four Bar Truss & Gear Train & Materials & Mic \\
\hline
MOEA/D-EGO & 7.052e-01(1.5e-02) & 6.227e-01(5.3e-02) & 7.658e-01(1.7e-02) & 5.365e-01(1.2e-01) & 1.164e+00(6.2e-03) & \cellcolor{gray!10}\textbf{1.247e+00(9.0e-03)} \\
TSEMO & \cellcolor{gray!50}\textbf{8.081e-01(5.8e-04)} & 7.117e-01(1.5e-02) & 8.520e-01(1.9e-03) & 7.811e-01(3.8e-02) & 1.155e+00(5.9e-03) & 1.238e+00(7.5e-03) \\
USEMO-EI & 7.882e-01(5.6e-03) & 7.057e-01(1.8e-02) & 8.287e-01(6.3e-03) & 6.742e-01(4.9e-02) & 1.150e+00(5.3e-03) & 1.243e+00(1.5e-02) \\
DGEMO & \cellcolor{gray!10}\textbf{8.068e-01(8.8e-04)} & 5.838e-01(5.9e-02) & 8.549e-01(3.0e-03) & \cellcolor{gray!10}\textbf{8.413e-01(6.8e-02)} & 1.162e+00(7.8e-03) & \cellcolor{gray!50}\textbf{1.272e+00(1.8e-02)} \\
qNEHVI & 7.990e-01(1.5e-03) & 7.105e-01(1.1e-02) & 8.240e-01(5.8e-03) & 7.810e-01(7.5e-02) & 1.150e+00(7.9e-03) & 1.232e+00(2.1e-02) \\
qParEGO & 7.744e-01(3.3e-03) & 6.852e-01(8.5e-03) & 7.651e-01(1.8e-02) & 6.832e-01(7.2e-02) & 1.142e+00(1.6e-02) & 1.176e+00(2.1e-02) \\
JES & 7.575e-01(5.5e-03) & 6.601e-01(1.6e-02) & 7.861e-01(9.1e-03) & 5.783e-01(7.0e-02) & 1.157e+00(4.3e-03) & 1.211e+00(2.1e-02) \\
\hline
FoMEMO-EI & 8.064e-01(1.2e-03) & \cellcolor{gray!50}\textbf{7.351e-01(8.0e-03)} & \cellcolor{gray!30}\textbf{8.656e-01(2.2e-03)} & \cellcolor{gray!30}\textbf{9.043e-01(7.1e-03)} & \cellcolor{gray!10}\textbf{1.171e+00(2.2e-03)} & 1.245e+00(1.4e-02) \\
FoMEMO-UCB & \cellcolor{gray!30}\textbf{8.072e-01(5.3e-04)} & \cellcolor{gray!30}\textbf{7.311e-01(5.1e-03)} & \cellcolor{gray!50}\textbf{8.698e-01(1.5e-03)} & \cellcolor{gray!50}\textbf{9.229e-01(5.3e-03)} & \cellcolor{gray!50}\textbf{1.176e+00(2.4e-03)} & 1.236e+00(1.9e-02) \\
FoMEMO-UHVI & 8.036e-01(6.4e-04) & \cellcolor{gray!10}\textbf{7.298e-01(6.5e-03)} & \cellcolor{gray!10}\textbf{8.567e-01(1.7e-03)} & 8.381e-01(1.2e-02) & \cellcolor{gray!30}\textbf{1.173e+00(1.5e-03)} & \cellcolor{gray!30}\textbf{1.250e+00(2.5e-02)} \\
FoMEMO-UR2I & 7.922e-01(1.8e-03) & 6.538e-01(9.8e-03) & 8.420e-01(4.5e-03) & 8.288e-01(2.1e-02) & 1.156e+00(4.5e-03) & 1.214e+00(1.2e-02) \\
\bottomrule[1pt]
\end{tabular}}
\end{table}

\begin{table}[htbp]
\centering
\caption{Means (stds) of hypervolume metrics ($\uparrow$) in parallel optimization ($q=10$) on representative problems over 10 independent runs. The best, second-best, and third-best mean results for each problem are highlighted with dark gray, gray, and light gray backgrounds, respectively.}\label{tab:q10 exp}
\renewcommand{\arraystretch}{1.5}
\scalebox{0.63}{
\begin{tabular}{ccccccc}
\toprule[1pt]
Algorithm & BC & ABD & Four Bar Truss & Gear Train & Materials & Mic \\
\hline
MOEA/D-EGO & 7.221e-01(2.1e-02) & 5.363e-01(6.9e-02) & 7.872e-01(2.0e-02) & 3.939e-01(1.3e-01) & 1.154e+00(5.7e-03) & 1.238e+00(1.1e-02) \\
TSEMO & \cellcolor{gray!50}\textbf{8.074e-01(7.6e-04)} & 6.512e-01(3.3e-02) & 8.521e-01(2.1e-03) & 7.060e-01(6.9e-02) & 1.145e+00(9.6e-03) & \cellcolor{gray!10}\textbf{1.242e+00(9.8e-03)} \\
USEMO-EI & 7.708e-01(1.4e-02) & 6.633e-01(2.2e-02) & 8.267e-01(1.2e-02) & 6.155e-01(9.7e-02) & 1.143e+00(7.9e-03) & 1.231e+00(1.4e-02) \\
DGEMO & \cellcolor{gray!10}\textbf{8.063e-01(1.3e-03)} & 5.228e-01(5.8e-02) & \cellcolor{gray!50}\textbf{8.713e-01(3.0e-03)} & 8.221e-01(5.7e-02) & 1.149e+00(7.3e-03) & \cellcolor{gray!10}\textbf{1.242e+00(1.3e-02)} \\
qNEHVI & 7.975e-01(2.4e-03) & \cellcolor{gray!10}\textbf{7.222e-01(7.5e-03)} & 8.194e-01(3.9e-03) & 7.600e-01(8.7e-02) & 1.145e+00(9.2e-03) & 1.207e+00(2.1e-02) \\
qParEGO & 7.722e-01(7.8e-03) & 6.892e-01(1.2e-02) & 7.286e-01(2.3e-02) & 6.207e-01(1.0e-01) & 1.150e+00(1.1e-02) & 1.175e+00(1.4e-02) \\
JES & 7.622e-01(3.2e-03) & 6.245e-01(2.1e-02) & 7.813e-01(8.3e-03) & 5.677e-01(6.8e-02) & 1.159e+00(3.0e-03) & 1.222e+00(1.6e-02) \\
\hline
FoMEMO-EI & \cellcolor{gray!10}\textbf{8.063e-01(8.5e-04)} & \cellcolor{gray!30}\textbf{7.248e-01(7.4e-03)} & \cellcolor{gray!10}\textbf{8.652e-01(1.8e-03)} & \cellcolor{gray!30}\textbf{8.987e-01(1.5e-02)} & \cellcolor{gray!10}\textbf{1.164e+00(4.7e-03)} & 1.237e+00(2.7e-02) \\
FoMEMO-UCB & \cellcolor{gray!30}\textbf{8.064e-01(6.5e-04)} & \cellcolor{gray!50}\textbf{7.347e-01(1.2e-02)} & \cellcolor{gray!30}\textbf{8.679e-01(2.8e-03)} & \cellcolor{gray!50}\textbf{9.188e-01(6.1e-03)} & \cellcolor{gray!50}\textbf{1.167e+00(5.4e-03)} & \cellcolor{gray!30}\textbf{1.247e+00(2.6e-02)} \\
FoMEMO-UHVI & 7.881e-01(2.1e-02) & 7.179e-01(1.0e-02) & 8.494e-01(1.7e-03) & \cellcolor{gray!10}\textbf{8.385e-01(1.6e-02)} & \cellcolor{gray!30}\textbf{1.166e+00(4.8e-03)} & \cellcolor{gray!50}\textbf{1.259e+00(1.5e-02)} \\
FoMEMO-UR2I & 7.297e-01(2.5e-02) & 5.862e-01(1.9e-02) & 7.577e-01(1.7e-02) & 6.736e-01(6.4e-02) & 1.137e+00(1.2e-02) & 1.180e+00(2.0e-02) \\
\bottomrule[1pt]
\end{tabular}}
\end{table}

\section{Acquisition Function Selection Guidelines}
\label{apd:acf_selection}
We first provide guidance on acquisition function selection based on the availability of user preference information. When the user specifies a preferred trade-off, FoMEMO-EI and FoMEMO-UCB are natural choices because they directly condition candidate generation on the preference to target the desired region of the Pareto front. Without a specified preference, all four variants are applicable. FoMEMO-EI and FoMEMO-UCB sample preferences across iterations, while FoMEMO-UHVI and FoMEMO-UR2I aggregate improvements across multiple preferences within each iteration.

The experimental results also provide practical guidance for selecting an appropriate acquisition function. We recommend FoMEMO-UCB as the default when no task-specific information is available. It provides the most consistent balance between optimization quality and query efficiency, particularly on engineering, many-objective, and parallel optimization tasks. FoMEMO-EI exhibits a similar performance profile in these settings but has lower query efficiency, making it a competitive alternative to FoMEMO-UCB. For HPO applications without a specified preference, FoMEMO-UHVI and FoMEMO-UR2I are also competitive options, as supported by their favorable experimental performance.

\section{Analysis of Method Applicability}
\label{apd:method applicability}
To generate synthetic data for model training, we employ a GP with an RBF kernel, which is known to be a universal kernel \cite{micchelli2006universal}. This choice allows the synthetic training distribution to cover a broad range of unknown functions. In principle, with sufficiently extensive training data from a universal kernel, the foundation model may encounter synthetic functions that closely resemble real objective functions, thereby enabling effective generalization. However, since it is impossible to train indefinitely to traverse the infinite space of functions and kernel hyperparameters, the model’s generalization ability is inevitably bounded by the prior. Therefore, investigating the scope of our methodology may benefit improvements to the current work and provide insights for future extensions.

Our exploration proceeds in two steps. First, from the perspective of methodology, we introduce a new model trained with mixed kernels, aiming to cover a broader family of GP priors. Second, from the perspective of test problems, we construct a counterexample to systematically illustrate when the method may fail and how such failures are related to prior mismatch.

\subsection{Training with Mixed Kernels}

\begin{table}[htbp]
\centering
\caption{Means of HV ($\Delta$ HV relative to RBF) over 10 independent runs. The better mean result for each algorithm compared to the base configuration is marked in dark green.}\label{tab:kernel ablation exp}
\renewcommand{\arraystretch}{1.5}
\scalebox{0.5}{
\begin{tabular}{clllllll}
\toprule
Model Configuration & Algorithm & BC & ABD & Four Bar Truss & Gear Train & Materials & Mic \\
\midrule

\multirow{4}{*}{Base} & FoMEMO-EI & 8.066e-01 & 7.308e-01 & 8.667e-01 & 9.107e-01 & 1.173e+00 & 1.261e+00 \\
& FoMEMO-UCB & 8.075e-01 & 7.369e-01 & 8.722e-01 & 9.270e-01 & 1.175e+00 & 1.262e+00 \\
& FoMEMO-UHVI & 8.052e-01 & 7.430e-01 & 8.609e-01 & 8.573e-01 & 1.176e+00 & 1.271e+00 \\
& FoMEMO-UR2I & 8.016e-01 & 7.126e-01 & 8.741e-01 & 8.961e-01 & 1.179e+00 & 1.274e+00 \\
\cmidrule(l){2-8}

\multirow{4}{*}{Mat\'ern-3/2 Kernel} & FoMEMO-EI & 8.071e-01 \textcolor{mydarkgreen}{(+5.0e-04)} & 7.354e-01 \textcolor{mydarkgreen}{(+4.6e-03)} & 8.676e-01 \textcolor{mydarkgreen}{(+9.0e-04)} & 9.134e-01 \textcolor{mydarkgreen}{(+2.7e-03)} & 1.175e+00 \textcolor{mydarkgreen}{(+2.0e-03)} & 1.263e+00 \textcolor{mydarkgreen}{(+2.0e-03)} \\
& FoMEMO-UCB & 8.063e-01 (-1.2e-03) & 7.501e-01 \textcolor{mydarkgreen}{(+1.3e-02)} & 8.693e-01 (-2.9e-03) & 9.145e-01 (-1.3e-02) & 1.175e+00 (+0.0e+00) & 1.235e+00 (-2.7e-02) \\
& FoMEMO-UHVI & 7.841e-01 (-2.1e-02) & 7.340e-01 (-9.0e-03) & 8.560e-01 (-4.9e-03) & 8.456e-01 (-1.2e-02) & 1.175e+00 (-1.0e-03) & 1.261e+00 (-1.0e-02) \\
& FoMEMO-UR2I & 8.018e-01 \textcolor{mydarkgreen}{(+2.0e-04)} & 7.192e-01 \textcolor{mydarkgreen}{(+6.6e-03)} & 8.697e-01 (-4.4e-03) & 8.890e-01 (-7.1e-03) & 1.174e+00 (-5.0e-03) & 1.244e+00 (-3.0e-02) \\
\cmidrule(l){2-8}

\multirow{4}{*}{Mat\'ern-5/2 Kernel} & FoMEMO-EI & 8.074e-01 \textcolor{mydarkgreen}{(+8.0e-04)} & 7.430e-01 \textcolor{mydarkgreen}{(+1.2e-02)} & 8.687e-01 \textcolor{mydarkgreen}{(+2.0e-03)} & 9.090e-01 (-1.7e-03) & 1.175e+00 \textcolor{mydarkgreen}{(+2.0e-03)} & 1.254e+00 (-7.0e-03) \\
& FoMEMO-UCB & 8.083e-01 \textcolor{mydarkgreen}{(+8.0e-04)} & 7.495e-01 \textcolor{mydarkgreen}{(+1.3e-02)} & 8.736e-01 \textcolor{mydarkgreen}{(+1.4e-03)} & 9.156e-01 (-1.1e-02) & 1.177e+00 \textcolor{mydarkgreen}{(+2.0e-03)} & 1.253e+00 (-9.0e-03) \\
& FoMEMO-UHVI & 8.074e-01 \textcolor{mydarkgreen}{(+2.2e-03)} & 7.396e-01 (-3.4e-03) & 8.610e-01 \textcolor{mydarkgreen}{(+1.0e-04)} & 8.383e-01 (-1.9e-02) & 1.175e+00 (-1.0e-03) & 1.267e+00 (-4.0e-03) \\
& FoMEMO-UR2I & 7.975e-01 (-4.1e-03) & 7.178e-01 \textcolor{mydarkgreen}{(+5.2e-03)} & 8.753e-01 \textcolor{mydarkgreen}{(+1.2e-03)} & 8.658e-01 (-3.0e-02) & 1.178e+00 (-1.0e-03) & 1.255e+00 (-1.9e-02) \\
\cmidrule(l){2-8}

\multirow{4}{*}{Mixed Kernel} & FoMEMO-EI & 8.062e-01 (-4.0e-04) & 7.356e-01 \textcolor{mydarkgreen}{(+4.8e-03)} & 8.680e-01 \textcolor{mydarkgreen}{(+1.3e-03)} & 9.159e-01 \textcolor{mydarkgreen}{(+5.2e-03)} & 1.173e+00 (+0.0e+00) & 1.242e+00 (-1.9e-02) \\
& FoMEMO-UCB & 8.077e-01 \textcolor{mydarkgreen}{(+2.0e-04)} & 7.392e-01 \textcolor{mydarkgreen}{(+2.3e-03)} & 8.714e-01 (-8.0e-04) & 9.147e-01 (-1.2e-02) & 1.177e+00 \textcolor{mydarkgreen}{(+2.0e-03)} & 1.248e+00 (-1.4e-02) \\
& FoMEMO-UHVI & 8.048e-01 (-4.0e-04) & 7.306e-01 (-1.2e-02) & 8.610e-01 \textcolor{mydarkgreen}{(+1.0e-04)} & 8.356e-01 (-2.2e-02) & 1.173e+00 (-3.0e-03) & 1.276e+00 \textcolor{mydarkgreen}{(+5.0e-03)} \\
& FoMEMO-UR2I & 8.017e-01 \textcolor{mydarkgreen}{(+1.0e-04)} & 7.331e-01 \textcolor{mydarkgreen}{(+2.0e-02)} & 8.733e-01 (-8.0e-04) & 8.705e-01 (-2.6e-02) & 1.167e+00 (-1.2e-02) & 1.266e+00 (-8.0e-03) \\
\bottomrule
\end{tabular}}
\end{table}

In this subsection, we train a new model using synthetic functions sampled equally from RBF, Mat\'ern-3/2, and Mat\'ern-5/2 kernels. The HV gaps relative to the RBF-trained model are reported in Table \ref{tab:kernel ablation exp}. The results show that simply broadening the training prior with mixed kernels does not consistently improve performance across all tasks and acquisition functions. This suggests that generalization is not determined only by the diversity of training kernels, but also by how well the kernel-induced function properties match the characteristics of the test problem, which we further examine in the next subsection.

\subsection{Interaction between Kernel Priors and Function Characteristics}

\begin{figure}[htbp]
\centering
\includegraphics[scale=0.52]{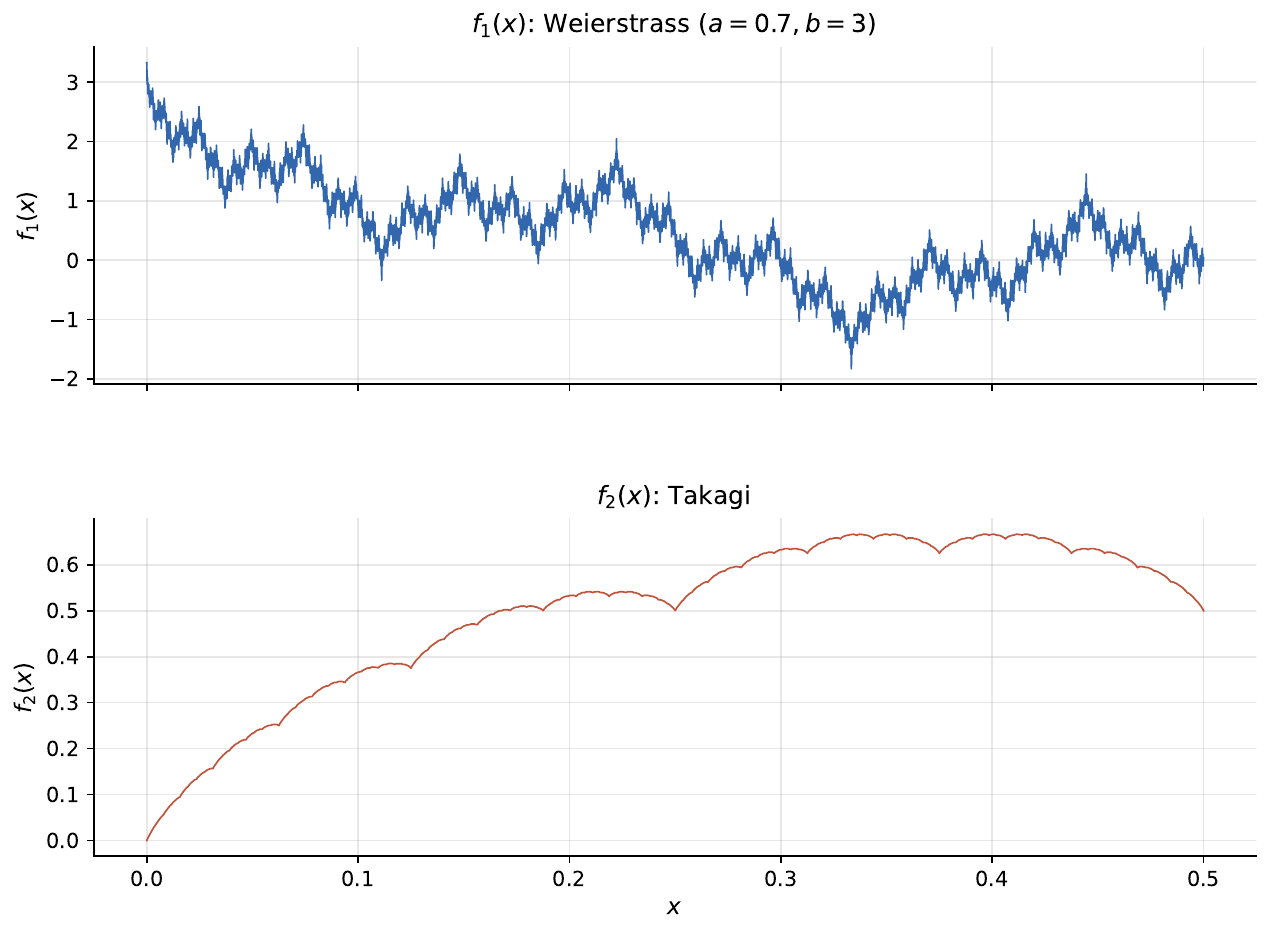}
\caption{Weierstrass-Takagi problem.}
\label{fig:WT problem}
\end{figure}

The mixed-kernel results above suggest that simply increasing the diversity of training kernels does not necessarily lead to uniformly better performance. To better understand when the training prior matters, we further construct a controlled extreme test problem using the Weierstrass\cite{hardy1916weierstrass} and Takagi \cite{takagi1990simple} functions as two objectives. Both functions are continuous but highly non-smooth, and therefore deliberately deviate from the strong smoothness prior induced by the RBF kernel, as illustrated in Figure \ref{fig:WT problem}. We restrict the domain to [0,0.5], where the two objectives exhibit a clear trade-off. The goal is not to propose a standard benchmark, but to create an interpretable example for analyzing how prior mismatch affects posterior prediction and downstream optimization.

\begin{table}[htbp]
\centering
\caption{Nominal H\"older regularity indicators for functions and kernels.}
\label{tab:holder exponent}
\renewcommand{\arraystretch}{1.3}
\scalebox{0.85}{
\begin{tabular}{cccccc}
\toprule[1pt]
Function / Kernel & Weierstrass & Takagi & Mat\'ern-3/2 & Mat\'ern-5/2 & RBF \\
\midrule
$\alpha$ & 0.325 & 1.0 & 1.5 & 2.5 & $\infty$ \\
\bottomrule[1pt]
\end{tabular}}
\end{table}

To characterize the roughness of these functions, we report their nominal H\"older regularity in Table \ref{tab:holder exponent}. The H\"older exponent $\alpha$ serves as an indicator of local regularity: smaller values correspond to rougher functions, while larger values indicate smoother behavior \cite{evans2022partial,gilbarg1998elliptic}. For the Weierstrass function, the exponent is given by $\alpha = -\frac{\log a}{\log b}$. With $a=0.7$ and $b=3$, this gives $\alpha \approx 0.325$, indicating a highly non-smooth objective. The Takagi function has critical H\"older regularity 1.0, and in our experiments we use its truncated numerical version. For GP kernels, the sample paths of a Mat\'ern-$\nu$ GP have local almost-H\"older regularity up to $\nu$. In contrast, the RBF kernel induces an infinitely smooth prior \cite{da2026sample}. Therefore, this problem is intentionally less aligned with the RBF training prior and more compatible with rougher Mat\'ern-type priors.

\begin{table}[htbp]
\centering
\caption{Regression metric means on Weierstrass-Takagi problem over 5 independent runs.}
\label{tab:WT regression}
\renewcommand{\arraystretch}{1.3}
\scalebox{0.78}{
\begin{tabular}{lccc}
\toprule
Kernel & RMSE ($\downarrow$) & MAE ($\downarrow$) & R2 ($\uparrow$) \\
\midrule
RBF      & 1.237e-02 & 9.588e-03 & 9.792e-01 \\
Mat\'ern-3/2 & \textbf{6.864e-03} & \textbf{5.125e-03} & \textbf{9.923e-01} \\
Mat\'ern-5/2 & 8.469e-03 & 6.334e-03 & 9.907e-01 \\
Mixed    & 7.265e-03 & 5.521e-03 & 9.917e-01 \\
\bottomrule
\end{tabular}}
\end{table}

Table \ref{tab:WT regression} compares the regression accuracy of models trained with different kernel priors on the Weierstrass-Takagi problem. The results show that Mat\'ern-based priors achieve lower prediction errors than the RBF prior, with Mat\'ern-3/2 performing best in terms of RMSE and MAE. This is consistent with the intended construction of the problem: the objectives are substantially less smooth than typical RBF sample paths, and therefore a rougher Mat\'ern prior better matches the function characteristics. The mixed-kernel model also improves over the RBF-trained model, but does not outperform the best single Mat\'ern prior, suggesting that simply increasing prior diversity is not always sufficient, the alignment between the training prior and the test-function characteristics remains important.

\begin{table}[htbp]
\centering
\caption{Means of HV ($\Delta$ HV relative to RBF) on Weierstrass-Takagi problem over 10 independent runs.}
\label{tab:WT optimization}
\renewcommand{\arraystretch}{1.3}
\scalebox{0.78}{
\begin{tabular}{lllll}
\toprule
Kernel & FoMEMO-EI & FoMEMO-UCB & FoMEMO-UHVI & FoMEMO-UR2I \\
\midrule
RBF      & 6.573e-01 & 6.469e-01 & 6.588e-01 & 6.323e-01 \\
Mat\'ern-3/2 & 6.674e-01 \textcolor{mydarkgreen}{(+1.0e-02)} & 6.548e-01 \textcolor{mydarkgreen}{(+7.9e-03)} & 6.768e-01 \textcolor{mydarkgreen}{(+1.8e-02)} & 6.599e-01 \textcolor{mydarkgreen}{(+2.8e-02)} \\
Mat\'ern-5/2 & 6.633e-01 \textcolor{mydarkgreen}{(+6.0e-03)} & 6.617e-01 \textcolor{mydarkgreen}{(+1.5e-02)} & 6.787e-01 \textcolor{mydarkgreen}{(+2.0e-02)} & 6.723e-01 \textcolor{mydarkgreen}{(+4.0e-02)} \\
Mixed    & 6.564e-01 (-9.0e-04) & 6.570e-01 \textcolor{mydarkgreen}{(+1.0e-02)} & 6.654e-01 \textcolor{mydarkgreen}{(+6.6e-03)} & 6.551e-01 \textcolor{mydarkgreen}{(+2.3e-02)} \\
\bottomrule
\end{tabular}}
\end{table}

The optimization results in Table \ref{tab:WT optimization} further support this observation. Mat\'ern-3/2 and Mat\'ern-5/2 generally improve the final HV over the RBF-trained model across the FoMEMO acquisition variants, while the mixed-kernel model also brings improvements in most cases. However, the mixed-kernel model does not uniformly dominate the single-kernel alternatives. These results suggest that the benefit of changing the training prior depends on whether the induced function regularity matches the target problem, rather than on kernel diversity alone. This provides a concrete example of how prior mismatch can affect both posterior prediction and downstream optimization performance.

\subsection{Summary and Future Directions}
\paragraph{Scope of This Work} In this paper, we focus on a general class of unconstrained expensive multi-objective optimization problems with continuous objective functions. Such problems are prevalent in numerous real-world applications and are often well suited for GP-based modeling, especially when the objectives exhibit certain smoothness and uncertainty quantification is important. This motivates our use of GP priors to generate large-scale synthetic training data, allowing the model to encounter a diverse set of possible objective landscapes that may exist in such real-world scenarios during pre-training. The experimental results demonstrate that this foundation-model-style paradigm can generalize effectively across a wide range of benchmark problems.

The analyses in this section further clarify the applicable scope of our method. Since the model is trained on synthetic functions sampled from specified priors, its generalization ability is inevitably influenced by how well these priors match the target problem. The mixed-kernel experiments show that increasing the diversity of training priors can be beneficial, but does not by itself guarantee uniformly better performance. From the perspective of function smoothness, the Weierstrass-Takagi test further illustrates that when the target objectives strongly deviate from the training prior, the prior mismatch can affect both posterior prediction and downstream optimization performance. These results suggest that prior diversity is useful, but the alignment between the training prior and the target function characteristics plays a more important role in determining generalization.

\paragraph{Potential Extensions} This observation also points to several directions for future work. First, expert knowledge can be incorporated to design more specialized priors during synthetic data generation, especially for domains with known structural properties such as non-smoothness, periodicity, constraints, or physical invariances. Second, instead of relying only on a fixed synthetic training distribution, the model could interact with the evaluation environment during optimization to identify potential prior mismatch and adaptively enrich the prior distribution. Third, rather than permanently freezing the model after a single round of synthetic pre-training, it may be useful to fine-tune the pre-trained model with a small amount of domain-specific data. Since synthetic data can be generated efficiently once a prior is specified, these extensions can be naturally integrated into our existing framework and may further improve its applicability to more specialized real-world optimization problems.%


\end{document}